\documentclass[11pt,a4paper]{lumia}

\usepackage[sort&compress]{natbib}
\usepackage{xspace}

\theoremstyle{plain}

\newtheorem*{proposition*}{Proposition}

\theoremstyle{definition}

\usepackage{graphicx}
\usepackage{tikz}
\usetikzlibrary{arrows.meta,positioning,fit,calc}
\usepackage{xurl}
\usepackage{xspace}
\usepackage{array}
\usepackage{boldline}
\usepackage{float}
\usepackage{multirow}
\usepackage{makecell}
\usepackage{ragged2e}
\usepackage{mathtools}
\usepackage{nicefrac}
\usepackage{subcaption}
\usepackage[export]{adjustbox}
\usepackage{afterpage}
\usepackage[capitalize,noabbrev]{cleveref}
\usepackage{fontawesome5}
\newcommand{\hflogo}{%
  \raisebox{-0.05\height}{%
    \includegraphics[height=1.00em]{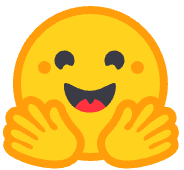}%
  }%
}

\newcolumntype{P}[1]{>{\RaggedRight\arraybackslash}p{#1}}
\newcolumntype{C}[1]{>{\Centering\arraybackslash}p{#1}}
\newcolumntype{M}[1]{>{\RaggedRight\arraybackslash}m{#1}}
\newcolumntype{N}[1]{>{\Centering\arraybackslash}m{#1}}
\newcolumntype{Y}[1]{>{\hsize=#1\hsize\linewidth=\hsize\RaggedRight\arraybackslash}X}
\newcolumntype{Z}[1]{>{\hsize=#1\hsize\linewidth=\hsize\Centering\arraybackslash}X}
\setlist[itemize]{leftmargin=18pt,noitemsep,topsep=2pt}
\setlist[enumerate]{leftmargin=20pt,noitemsep,topsep=2pt}

\definecolor{originaccent}{HTML}{8B6B3F}
\colorlet{lumiaabsbg}{originaccent!10!white}
\colorlet{origintablehead}{originaccent!20!white}
\colorlet{origintablerow}{black!4}
\definecolor{originblue}{HTML}{245E7A}
\definecolor{originlight}{HTML}{EEF4F7}
\definecolor{verifyred}{HTML}{9C2F2F}
\renewcommand{\titlefont}{\color{black}\normalfont\bfseries\fontsize{21}{23}\selectfont}
\newcommand{\project}{SolarWM\xspace}

\newcommand{\pubyes}{\ding{51}}
\newcommand{\pubpartial}{\ensuremath{\triangle}}

\newcommand{\pubno}{\textemdash}

\setheadertext{\includegraphics[height=0.8em]{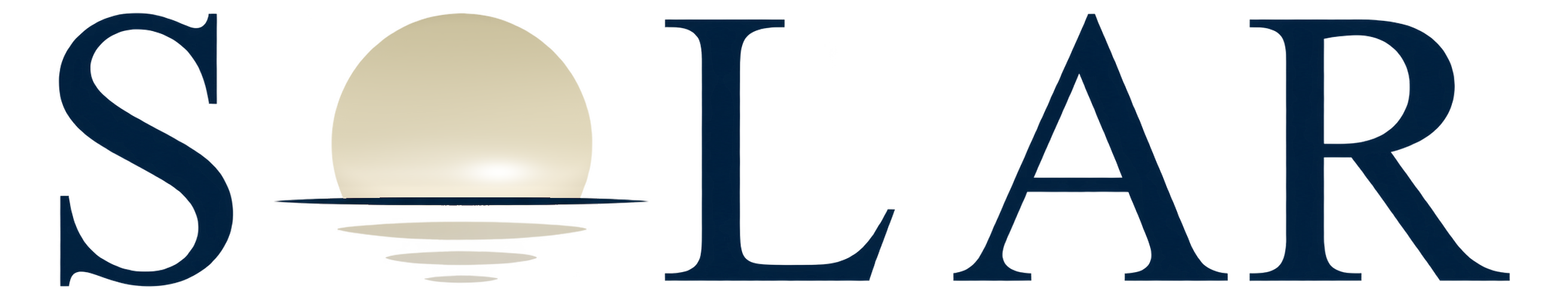}}
\setreportdate{2026.9}
\title{\vspace{-1mm}
  \textcolor{originaccent}{\project:}
  Open Data and Scalable Training for Long-Horizon Video World Models
}

\setheadertitle{\includegraphics[height=0.8em]{imgs/solar-logo-transparent.png}}

\author{
\mbox{Junchao Huang\textsuperscript{1,2}},\hspace{0.3em}
\mbox{Guian Fang\textsuperscript{3}},\hspace{0.3em}
\mbox{Shengju Qian\textsuperscript{4,$\dagger$}},\hspace{0.3em}
\mbox{Xianghao Kong\textsuperscript{5}},\hspace{0.3em}
\mbox{Zhuoran Zhao\textsuperscript{5,6}}
\par
\vspace{-6pt}

\mbox{Wei Huang\textsuperscript{7}},\hspace{0.32em}
\mbox{Yihua Du\textsuperscript{6}},\hspace{0.32em}
\mbox{Zixin Zhang\textsuperscript{6}},\hspace{0.32em}
\mbox{Justin Cui\textsuperscript{8}},\hspace{0.32em}
\mbox{Yuchao Gu\textsuperscript{7}},\hspace{0.32em}
\mbox{Yukang Chen\textsuperscript{7}},\hspace{0.32em}
\mbox{Xinting Hu}
\par
\vspace{-6pt}

\mbox{Tianyu He\textsuperscript{9}},\hspace{0.3em}
\mbox{Shaoshuai Shi},\hspace{0.3em}
\mbox{Zhuotao Tian\textsuperscript{2}},\hspace{0.3em}
\mbox{Xin Wang},\hspace{0.3em}
\mbox{Mike Zheng Shou\textsuperscript{3}},\hspace{0.3em}
\mbox{Li Jiang\textsuperscript{1,2,$\ddagger$}}
\par
\vspace{-4pt}

{\normalfont\fontsize{8}{10}\selectfont
\textsuperscript{1}CUHK-SZ
\quad
\textsuperscript{2}SLAI
\quad
\textsuperscript{3}NUS
\quad
\textsuperscript{4}CUHK
\quad
\textsuperscript{5}HKUST
\quad
\textsuperscript{6}HKUST-GZ
\quad
\textsuperscript{7}NVIDIA
\quad
\textsuperscript{8}UCLA
\quad
\textsuperscript{9}MSRA
}
\par\vspace{-6pt}
}

\correspondingemail{%
\emailicon\ 
\textcolor{black}{\textnormal{junchaoh.cs@gmail.com}}
\qquad
\href{https://huggingface.co/datasets/junchaoh-cs/SolarWM-Data}{
    \hflogo\, \textcolor{black}{Dataset}
}
\qquad
\href{https://github.com/Junchao-cs/SolarWM/tree/main}{
    \textcolor{black}{{\faGithub}\, Code}
}
\qquad
$^\dagger$ Project Lead
\qquad
$^\ddagger$ Corresponding Author
\vspace{-1mm}
}

\makeatletter
\let\origin@oldmaketitle\maketitle
\renewcommand{\maketitle}{%
  \origin@oldmaketitle
  \par\vspace{-0mm}%
  \begingroup
    \centering
    \includegraphics[width=0.95\textwidth]{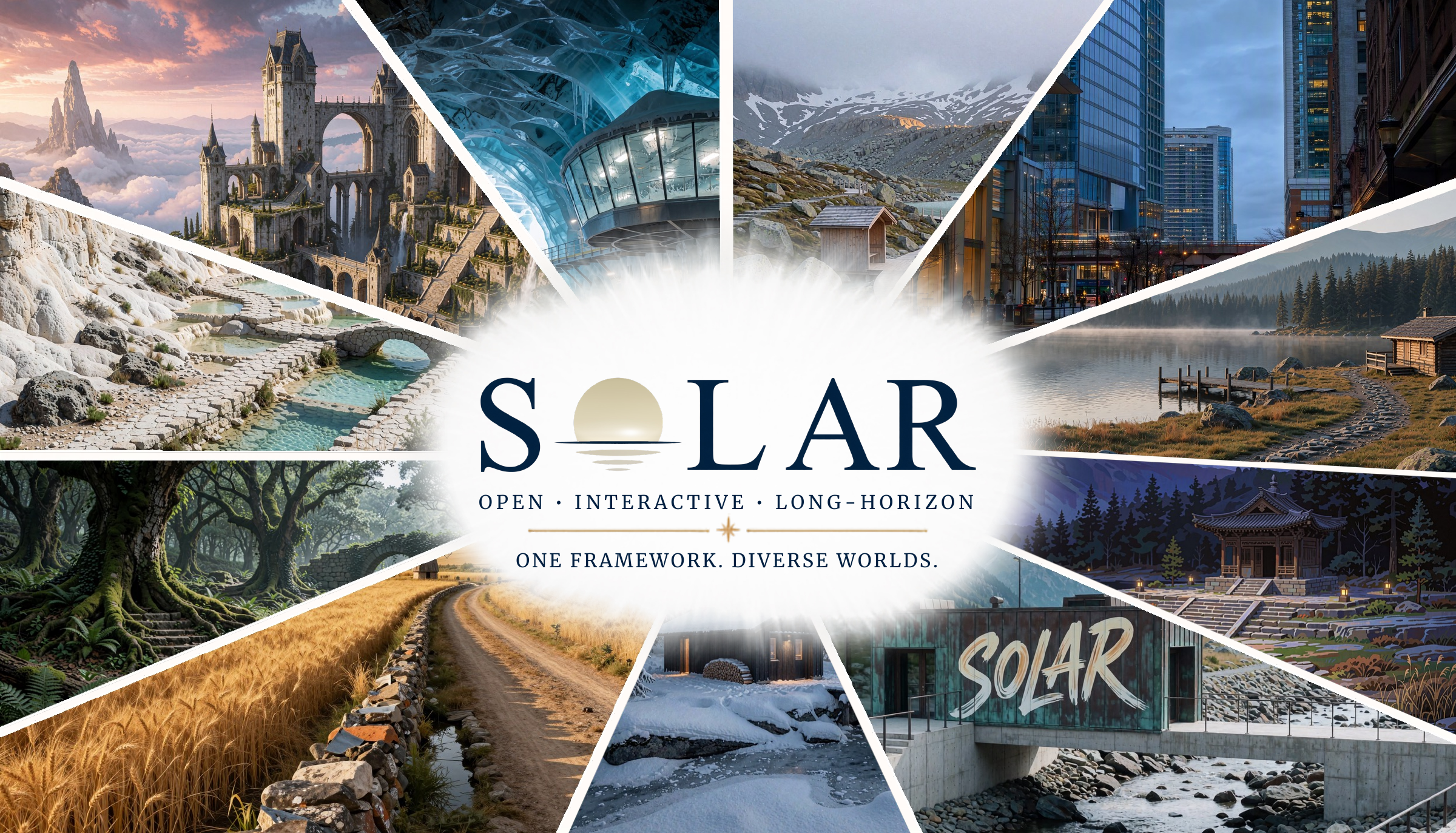}\par
    \vspace{0mm}%
    \captionsetup{type=figure}%
    \caption{\textbf{\project: an open foundation for interactive, long-horizon video world models.}  
    }%
    \label{fig:teaser}%
    \par
  \endgroup
  \vspace{0mm}%
}
\makeatother

\begin{document}
\begin{abstract}
We introduce \project, a fully open foundation for building interactive video world models from data preparation through long-horizon inference.
Training across heterogeneous data sources and video backbones is challenging: datasets differ in temporal scale, camera geometry, visual quality, motion, and captioning styles, while video generators use distinct representations and architectures. Naive data mixing and model-specific implementations therefore produce inconsistent supervision and make results difficult to reproduce and compare.
\project addresses this coupling with a reconfigurable multi-source data engine and a backbone-native adaptation framework. The engine converts 1.43 million canonical clips from 10 datasets into a unified, frame-aligned contract covering visual observations, metric camera geometry, captions, quality metadata, selection decisions, and provenance, while decoupling source processing from mixture construction. Under shared camera-conditioning, training, and inference interfaces, we instantiate four 5B--33B models based on Wan2.2, LTX-2.5, and MiniMax-H3 while preserving their native representations and objectives. A unified three-stage recipe combines bidirectional adaptation, teacher-forced autoregressive initialization, and distribution matching distillation. The resulting causal models enable real-time interaction over rollouts ranging from minutes to hours after being trained on only 5s sequences.
By releasing the resulting data, pipeline, recipes, weights, and framework, \project provides a reproducible and extensible foundation for interactive world-model research.

\par\medskip
\vspace{-1mm}
\noindent\textbf{Website (Dataset \& Code \& Model):}
{\fontsize{8.8}{10.5}\selectfont
\href{https://junchao-cs.github.io/SolarWM-Web/}
{\nolinkurl{https://junchao-cs.github.io/SolarWM-Web/}}
}

\end{abstract}

\maketitle

\section{Introduction}
\label{sec:introduction}

\begin{quote}
 \ \ \ \ \ \ \ \ \ \ \emph{``Imagination will often carry us to worlds that never were.''} \ \ \ \ --- Carl Sagan
\end{quote}

Interactive video world models imagine future visual observations conditioned on camera motion, player actions, or semantic instructions, transforming passive video generators into environments that can be interactively explored and controlled
\citep{bruce2024genie,valevski2024gamengen,mao2025yume,
zhang2026robostereo}. Recent systems have demonstrated increasingly realistic interaction in game environments and open visual domains, highlighting their potential for simulation, embodied learning, and interactive content creation \citep{he2025matrixgame2,dreamxteam2026dreamx,
zhu2026sanawm,huang2025edit360,wang2026liveedit,gao2026lingbot2,zhang2026symphomotion,kong2026bico}.
However, the transition from short-clip generation to interactive long-horizon rollout requires models to maintain visual quality and temporal coherence while remaining responsive to control
\citep{wang2026matrixgame3,alayaworld2026v11,jiang2026abotworld,gao2026lingbot2}.
 Achieving this capability requires coherent supervision
across heterogeneous data sources and effective adaptation of diverse video-generation backbones without compromising their pretrained capabilities.

These two requirements are closely coupled. On the data side, existing datasets differ in temporal scale, visual quality, motion distribution, captioning styles, and camera conventions
\citep{ju2024miradata,ling2023dl3dv,zheng2025realcam,
wang2025spatialvid,li2025sekai,hu2026physeditworld}. Naively combining them can introduce inconsistent supervision, causing models that perform well on individual sources to deteriorate under multi-source training. On the model side, video generators differ substantially in architecture, including their latent representations, attention structures, and conditioning mechanisms
\citep{zhao2026minwm,rui2026biwm}. Although a shared adaptation strategy is desirable for scalability, overlooking this heterogeneity can compromise the pretrained capabilities; conversely, backbone-specific adaptation limits systematic comparison and extensibility.
Existing open systems often overlook this coupling, relying on restricted data sources and largely model-specific implementations \citep{zhu2026sanawm,zhao2026minwm,rui2026biwm,li2026forgewm}.
Consequently, the field still lacks a unified and reproducible foundation that coordinates multi-source data construction and backbone adaptation while providing a standardized and efficient training protocol across backbones.

We introduce \project, a fully open and unified interactive video world-model foundation comprising a reconfigurable multi-source data infrastructure and a scalable backbone-native adaptation framework. At the data level, \project converts
heterogeneous sources into a common training contract with consistent temporal, geometric, semantic, and quality supervision. At the model level, \project provides shared interfaces for camera conditioning, optimization, and rollout
while preserving the native representations of individual video generators. We further establish a unified three-stage training recipe for heterogeneous backbones that combines simplicity with high efficiency, comprising bidirectional model training, autoregressive (AR) adaptation \citep{chen2024diffusion,gu2026anyflowanystepvideodiffusion}, and distribution matching distillation (DMD) \citep{yin2024one,huang2026self,zhuang2026selfgradientforcing}.

The data infrastructure contains approximately 1.43 million canonical clips from 10 source datasets, corresponding to over 25 TB of physical data. The corpus spans real-world, synthetic, and game environments across diverse scene layouts, camera motions, temporal scales, and motion patterns. Each source is converted into a canonical representation that aligns visual observations, camera geometry, language supervision, and quality metadata. We release the complete processed corpus together with the end-to-end processing pipeline, enabling reproducible reconstruction, extension, and reuse.
A key design principle of the infrastructure is to construct a reconfigurable and comprehensive training resource for world models by decoupling computationally intensive source preprocessing from training-mixture construction. Researchers can reproduce the released mixture or define alternative recipes by modifying filtering criteria, sampling ratios, and source weights without rerunning source-level processing.

Built on the canonical corpus produced by this infrastructure, we instantiate a scalable \project model family comprising \project-wan2.2-5B, \project-wan2.2-14B, \project-ltx-2.5-22B, and \project-minimax-h3-33B \citep{wan2025,hacohen2025ltx2,minimax2026h3}, with model sizes spanning 5B to 33B parameters. The backbone-native adaptation introduces only the interfaces required for camera conditioning and long-horizon rollout while preserving each generator's native representations and optimization objectives. This design yields an extensible and directly comparable model family that supports high-fidelity, camera-controllable video generation across real-world, synthetic, and game environments.

Experiments further reveal that strong long-horizon world models can be obtained with a simple yet highly efficient training procedure.
First, our models achieve state-of-the-art performance without specialized ODE or consistency-distillation (CD) initialization \citep{yin2025slow,huang2026self,zhu2026causal}. Second, most optimization should be performed during bidirectional training; AR adaptation then converges rapidly, while DMD requires even fewer optimization steps. Third, after training on only \textit{5s} sequences, the resulting models support open-ended rollouts over \textit{minutes-to-hours} horizons without additional long-sequence fine-tuning or attention-sink mechanisms. These key findings show that scalable long-horizon generation can be achieved without a complex training pipeline or computationally intensive training on long videos, provided that data construction, backbone adaptation, and training stages are properly aligned.

Our main contributions are summarized as follows:
\begin{itemize}
\item \textbf{A fully open and unified foundation.}
We introduce \project, a fully open and unified interactive video world-model foundation integrating a reconfigurable multi-source data infrastructure with a scalable backbone-native adaptation framework across data construction, model
training, and long-horizon inference.
\vspace{1mm}
\item \textbf{A reconfigurable multi-source data infrastructure.}
We process approximately 1.43 million clips from 10 datasets covering real-world, synthetic, and game environments, and release the corpus and processing pipeline to support reproducible and flexible training-mixture
construction.
\vspace{1mm}
\item \textbf{A scalable backbone-native adaptation framework and model family.}
We train four 5B--33B models across heterogeneous backbones using a unified data recipe and a simple, efficient three-stage training procedure comprising bidirectional training, AR adaptation, and DMD. The models achieve state-of-the-art performance and support open-ended rollouts over \textit{minutes-to-hours} horizons after training solely on \textit{5s} sequences.
\end{itemize}

\section{Open-Source Release}
\label{sec:release}

We will release the complete processed corpus including all 1.43 million canonical clips from 10 source datasets, together with their annotations, metadata, intermediate records, quality assessments, selection decisions, and provenance information, as well as the end-to-end data-processing pipeline. Because source-level preprocessing is decoupled from mixture construction, researchers can define alternative training recipes by changing filtering criteria, sampling ratios, source weights, or split definitions, without rerunning the computationally intensive source preprocessing. We will also release the training recipes, model weights, and implementations of bidirectional training, AR adaptation and rollout, and DMD training and inference. Collectively, these artifacts provide a complete and reproducible foundation for interactive video world-model research, enabling flexible data-mixture design, controlled cross-backbone comparison, and systematic extension to new data and models.

\section{Related Work}
\label{sec:related}

\subsection{Interactive Video World Models}
\label{sec:related_models}

Interactive video world models extend passive video generation into visual
simulation: given an initial observation, they roll out future observations in
response to actions, camera motion, or semantic instructions. Early systems
focused on compact action spaces: Genie learns latent actions from Internet
videos \citep{bruce2024genie}, DIAMOND models Atari environments
\citep{alonso2024diamond}, and GameNGen simulates \textit{DOOM} from recorded
state--action trajectories \citep{valevski2024gamengen}. Minecraft later became
a common testbed for autoregressive and streaming interaction, as illustrated
by MineWorld \citep{guo2025mineworld}, Memory Forcing \citep{huang2025memoryforcingspatiotemporalmemory}, and the Matrix-Game series
\citep{zhang2025matrixgame,he2025matrixgame2}. Their progress
establishes video generators
as interactive simulators, but many remain tied to a particular environment,
collection process, or control vocabulary.

Recent work extends the setting to open-domain visual exploration. Yume,
WorldPlay, and AlayaWorld support camera- or keyboard-controlled world
extension with mechanisms for long-range consistency
\citep{mao2025yume,sun2025worldplay,alayaworld2026v11}. LingBot-World,
ABot-World-0, DreamX-World, SANA-WM, Genie~3, LIVE, and recent Matrix-Game variants
further improve visual quality, rollout length, interaction, or deployment
efficiency
\citep{robbyant2026lingbot,gao2026lingbot2,jiang2026abotworld,
dreamxteam2026dreamx,zhu2026sanawm,parkerholder2025genie3,
wang2026matrixgame3,huang2026livelonghorizoninteractivevideo,riemanndynamics2026matrixgame35}. Together, these systems
show that an interactive world model depends not only on the base video generation model, but also on the data, adaptation, causal-training, and inference pipelines surrounding
it.

\subsection{Training Frameworks and Open Research Stacks}
\label{sec:related_training}

Most high-quality video generators are pretrained with bidirectional temporal
attention, whereas interactive rollout requires causal prediction from
generated history. Diffusion Forcing provides a causal sequence formulation
\citep{chen2024diffusion}; later methods convert bidirectional video models
through staged distillation, train on model-generated context to reduce
exposure bias, and enable few-step sampling through ODE, consistency, flow-map,
or distribution-matching objectives
\citep{yin2025slow,huang2026self,zhu2026causal,
zhao2026causalforcingscalablefewstep,gu2026anyflowanystepvideodiffusion,
yin2024one}. Astrolabe studies forward-process reinforcement learning
for post-training distilled autoregressive video models
\citep{zhang2026astrolabe}. KVPO further proposes ODE-native GRPO
for aligning streaming autoregressive video models through semantic
exploration over historical KV caches \citep{zhang2026kvpo}. These methods provide the main ingredients for long-horizon causal
generation, but they do not by themselves provide a unified training stack
that can be applied reproducibly across heterogeneous video backbones.

Beyond these algorithmic advances, only a limited number of systems release
training implementations. DIAMOND, Yume, minWM, BiWM, ForgeWM, SANA-WM,
WorldPlay, and AlayaWorld expose code for some or all training stages
\citep{alonso2024diamond,mao2025yume,zhao2026minwm,rui2026biwm,
li2026forgewm,zhu2026sanawm,sun2025worldplay,alayaworld2026v11}, whereas most
systems release only checkpoints, inference code, or demonstrations. Even when
training code is available, the released package is not always directly
reproducible or reconfigurable: processed data, source-to-training
construction, exact selection and mixture recipes, or checkpoint-matched
optimization configurations may still be missing. Reproducing a model or
adapting the pipeline to a new dataset or backbone can therefore require
reconstructing substantial parts of the original system. Moreover, existing
multi-backbone releases typically cover only two or three backbone families,
leaving it unclear whether a training design transfers broadly or depends on
model-specific engineering. \Cref{tab:world-model-openness} provides a systematic comparison of
representative interactive video world models based on their official project
pages, code repositories, and released artifacts. Specifically, it examines
the availability of model weights, inference and training code, processed
training data, complete data-selection records, executable source-to-training
pipelines, exact data and optimization recipes, and support for multiple
video-backbone families
\citep{alonso2024diamondrepo,decart2024oasisrepo,
microsoft2026mineworldrepo,skywork2026matrixgamerepo,
robbyant2026lingbotrepo,robbyant2026lingbot2repo,
amap2026abotworldrepo,parkerholder2025genie3,
dreamxteam2026dreamxrepo,nvlabs2026sanawmrepo,yume2026repo,
tencent2026worldplayrepo,shengshu2026minwmrepo,lynnreal2026biwmrepo,
alayalab2026alayaworldrepo,forgewm2026repo,
riemanndynamics2026matrixgame35repo}.

\project addresses both gaps with a compact, unified, and fully open training
framework. The same three-stage recipe---bidirectional camera-conditioned
adaptation, teacher-forced AnyFlow autoregressive initialization, and DMD-based
causal training---is designed for all four models in the \project family:
\project-wan2.2-5B, \project-wan2.2-14B, \project-ltx-2.5-22B, and
\project-minimax-h3-33B. Rather than assembling a separate training system for
each model, \project shares the data, camera-conditioning, optimization, and
rollout interfaces while retaining only the backbone-native representations
and objectives required by each generator. This demonstrates that the simple
recipe is a reusable cross-backbone framework, rather than a model-specific
training procedure.

We will open-source the complete stack for all four model instantiations,
including the implementation and configuration of every training stage,
backbone adapters, processed corpus and complete selection records, executable
source-to-training pipeline, exact data-mixture and optimization recipes,
checkpoint-matched settings, model weights, and inference code. The released
artifacts will therefore support both end-to-end reproduction of each of the
four models and controlled reconfiguration across data filters, mixtures,
backbones, schedules, and stage settings. The value of \project lies not only
in simplifying causal world-model training, but also in making a
four-model, recipe-complete research stack available for direct reproduction,
comparison, and extension.

\begin{table}[t]
\centering
\caption{\textbf{Release matrix for representative interactive video world models.}
Release status was verified from official artifacts as of August 18, 2026; \project entries indicate commitments for this release.
\pubyes: public; \pubpartial: reduced, or temporarily unavailable; \pubno: not identified.
``Pipeline'': executable source-to-training workflow; ``Full Rec.'': complete annotations, scores, provenance, and decisions; ``Exact Recipe'': executable data and optimization configurations.
Statistics cover public or committed payloads; ``Multi-BB'' denotes multiple video-backbone families.}
\vspace{-0.8em}
\label{tab:world-model-openness}
\renewcommand{\arraystretch}{1.25}
\resizebox{\textwidth}{!}{%
\begin{tabular}{l|>{\centering\arraybackslash}m{3.2cm}ccc>{\centering\arraybackslash}m{4.5cm}cccc>{\centering\arraybackslash}m{2.1cm}}
\hlineB{2.5}
\rowcolor{origintablehead}
\textbf{System} & \textbf{Domain} & \textbf{Weights} &
\textbf{Infer.} & \textbf{Train} & \textbf{Data} & \textbf{Full Rec.} &
\textbf{Pipeline} & \textbf{Exact Recipe} & \textbf{Multi-BB} & \textbf{Link} \\
\hlineB{1.5}
\multicolumn{11}{c}{\textit{Game- and action-centric systems}} \\
\hline
\rowcolor{origintablerow}
DIAMOND & Atari / CSGO
  & \pubyes & \pubyes & \pubyes & \pubno & \pubno & \pubno & \pubyes & \pubno
  & \href{https://diamond-wm.github.io/}{Project} \\
Oasis & Minecraft
  & \pubpartial & \pubyes & \pubno & \pubno & \pubno & \pubno & \pubno & \pubno
  & \href{https://oasis-model.github.io/}{Project} \\
\rowcolor{origintablerow}
MineWorld & Minecraft
  & \pubpartial & \pubyes & \pubno & \pubno & \pubno & \pubno & \pubno & \pubno
  & \href{https://github.com/microsoft/mineworld}{GitHub} \\
Matrix-Game 2.0 & Game / synthetic
  & \pubyes & \pubyes & \pubno & \pubno & \pubno & \pubno & \pubno & \pubno
  & \href{https://matrix-game-v2.github.io/}{Project} \\
\rowcolor{origintablerow}
Matrix-Game 3.0 & Game / real-world
  & \pubpartial & \pubyes & \pubno & \pubno & \pubno & \pubno & \pubno & \pubno
  & \href{https://matrix-game-v3.github.io/}{Project} \\
ABot-World-0 & Multi-source
  & \pubpartial & \pubyes & \pubno
  & \makecell{\pubyes~2.74 TB, 30k clips}
  & \pubpartial & \pubno & \pubno & \pubno
  & \href{https://amap-cvlab.github.io/ABot-World/}{Project} \\
\rowcolor{origintablerow}
ForgeWM & Game
  & \pubyes & \pubyes & \pubyes
  & \makecell{\pubyes~95 GB, 40k clips}
  & \pubno & \pubyes & \pubyes & \pubno
  & \href{https://github.com/asdfo123/ForgeWM}{GitHub} \\
\hline
\multicolumn{11}{c}{\textit{Multi-domain and general-purpose systems}} \\
\hline
\rowcolor{origintablerow}
Google Genie 3 & Multi-domain
  & \pubno & \pubno & \pubno & \pubno & \pubno & \pubno & \pubno & \pubno
  & \href{https://deepmind.google/models/genie/}{Project} \\
LingBot-World & Multi-domain
  & \pubyes & \pubyes & \pubno & \pubno & \pubno & \pubno & \pubno & \pubno
  & \href{https://technology.robbyant.com/lingbot-world}{Project} \\
\rowcolor{origintablerow}
LingBot-World 2.0 & Multi-domain
  & \pubpartial & \pubyes & \pubno & \pubno & \pubno & \pubno & \pubno & \pubno
  & \href{https://technology.robbyant.com/lingbot-world-v2}{Project} \\
Matrix-Game 3.5 & Multi-domain
  & \pubyes & \pubyes & \pubno & \pubno & \pubno & \pubno & \pubno & \pubno
  & \href{https://github.com/Riemann-Dynamics/Matrix-Game-3.5}{GitHub} \\
\rowcolor{origintablerow}
Yume & Multi-domain
  & \pubyes & \pubyes & \pubyes & \pubno & \pubno & \pubpartial & \pubpartial & \pubno
  & \href{https://github.com/stdstu12/YUME}{GitHub} \\
DreamX-World 1.0 & Multi-domain
  & \pubpartial & \pubyes & \pubno & \pubno & \pubno & \pubno & \pubno & \pubno
  & \href{https://amap-ml.github.io/DreamX_World/}{Project} \\
\rowcolor{origintablerow}
SANA-WM & Multi-domain
  & \pubpartial & \pubyes & \pubyes
  & \makecell{\pubpartial~235 GB, 1.6k clips}
  & \pubpartial & \pubpartial & \pubpartial & \pubno
  & \href{https://nvlabs.github.io/Sana/WM/}{Project} \\
HY-WorldPlay 1.5 & Multi-domain
  & \pubpartial & \pubyes & \pubyes & \pubno & \pubno & \pubpartial & \pubpartial & \pubyes~(2)
  & \href{https://github.com/Tencent-Hunyuan/HY-WorldPlay}{GitHub} \\
\rowcolor{origintablerow}
minWM & Multi-domain
  & \pubyes & \pubyes & \pubyes
  & \makecell{\pubpartial~527 GB, 19k clips}
  & \pubno & \pubyes & \pubyes & \pubyes~(2)
  & \href{https://github.com/shengshu-ai/minWM}{GitHub} \\
BiWM & Multi-domain
  & \pubno & \pubyes & \pubyes
  & \makecell{\pubyes~16.9 GB, 25k clips}
  & \pubno & \pubpartial & \pubpartial & \pubyes~(3)
  & \href{https://github.com/LynnReal-AI/BiWM}{GitHub} \\
\rowcolor{origintablerow}
AlayaWorld v1.1 & Multi-domain
  & \pubyes & \pubyes & \pubyes
  & \makecell{\pubpartial~1.5 GB, 18k clips}
  & \pubno & \pubpartial & \pubpartial & \pubno
  & \href{https://github.com/AlayaLab/AlayaWorld}{GitHub} \\
\hline
\rowcolor{originaccent!12!white}
\textbf{\project (Ours)} & \textbf{Multi-domain}
  & \pubyes & \pubyes & \pubyes
  & \makecell{\pubyes~25.85 TB, 1426k clips}
  & \pubyes & \pubyes & \pubyes
  & \pubyes(4)
  & \href{https://junchao-cs.github.io/SolarWM-Demo/}{Project} \\
\hlineB{2.5}
\end{tabular}%
}
\vspace{-3mm}
\end{table}

\subsection{Data for Interactive World Modeling}
\label{sec:related_data}

Data for interactive world modeling draws on several overlapping data
traditions. Large video--text corpora such as WebVid-10M
\citep{bain2021frozen}, InternVid \citep{wang2023internvid}, and MiraData
\citep{ju2024miradata} provide visual and semantic diversity but generally lack
frame-aligned geometric control. Camera-oriented datasets such as DL3DV-10K,
RealCam-Vid, SpatialVID, and MultiCamVideo provide complementary pose, depth,
or trajectory supervision
\citep{ling2023dl3dv,zheng2025realcam,wang2025spatialvid,
bai2025recammaster}. Multi-domain collections including OmniWorld, Sekai,
MIND, and ABOT-World broaden the range of scenes and interaction settings
\citep{zhou2025omniworld,li2025sekai,ye2026mind,jiang2026abotworld}.

These sources are complementary but not directly interchangeable. They differ
in temporal scale, resolution, visual quality, camera evidence, coordinate
conventions, metric scale, captions, and motion distributions. Their released
media and annotations also do not necessarily specify the processing that
produces model-ready samples. The challenge is therefore to construct a
consistent and auditable training contract, not simply to concatenate more
datasets.

\project addresses this gap with the open data engine in \cref{sec:data}. It
normalizes heterogeneous sources into a frame-aligned representation and
separates source processing from mixture construction. We will release the
processed corpus, annotations, quality and provenance records, accepted and
rejected samples, executable pipeline, and exact recipes. This allows users to
reproduce our mixtures or reconfigure filters, source weights, temporal views,
and model readers without rerunning the annotation pipeline.

\section{Training Pipeline}
\label{sec:training}

\begin{figure*}[t]
  \centering
  \includegraphics[width=\linewidth]{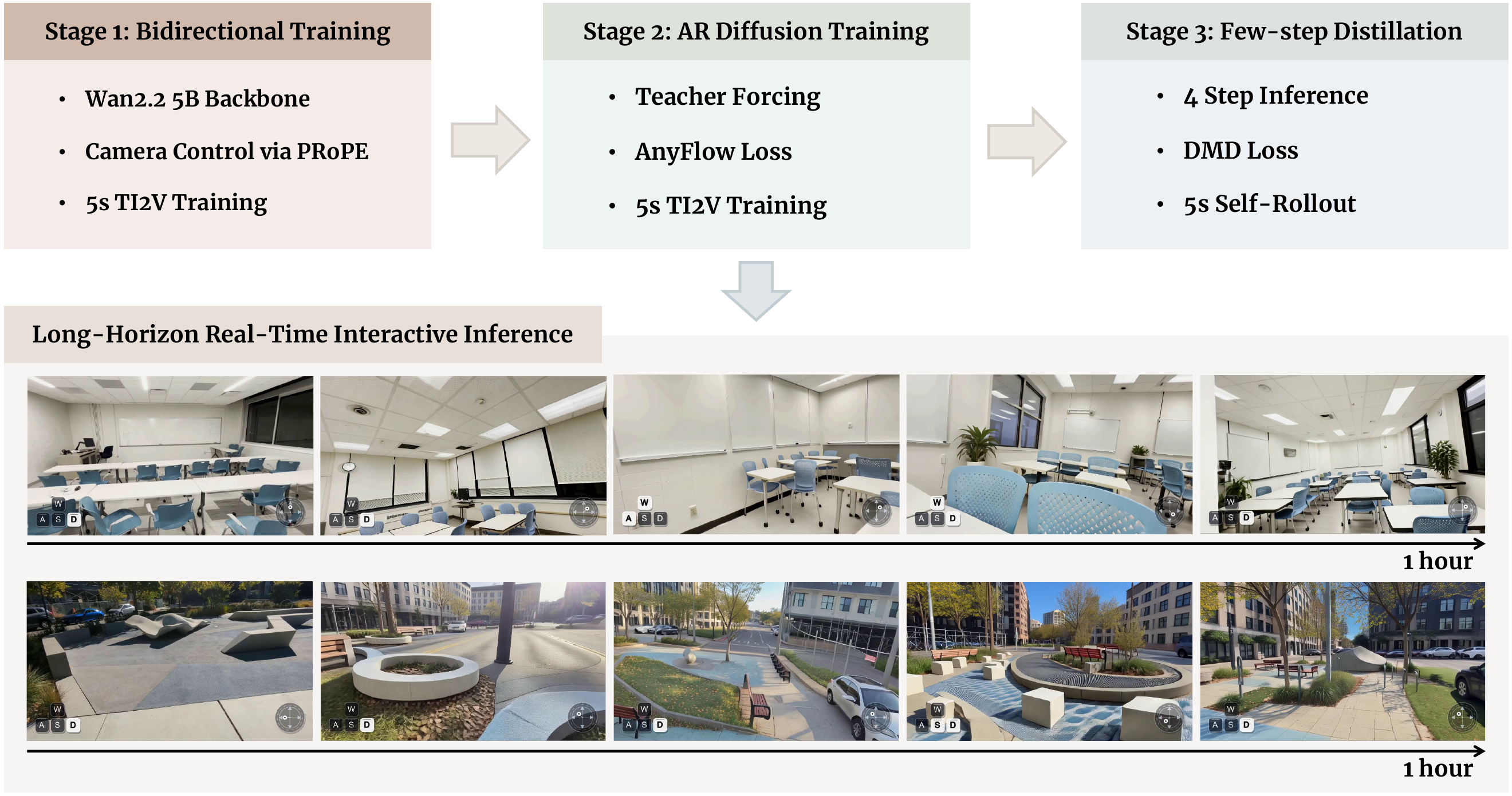}
  \caption{\textbf{Overview of the SolarWM-5B training pipeline and hour-scale inference results}}
  \label{fig:pipeline}
\end{figure*}

We train the world model in three stages: bidirectional adaptation, teacher-forced AnyFlow autoregressive initialization, and DMD-based causal training. The first stage adapts the pretrained video generator to camera-conditioned world data under bidirectional attention. The second switches to causal attention and combines teacher forcing with the AnyFlow loss \citep{gu2026anyflowanystepvideodiffusion}, directly producing a few-step autoregressive initializer. This stage is fast because it mainly activates causal prediction in the well-trained bidirectional model rather than relearning appearance and motion. By making the autoregressive initializer few-step from the outset, TF-AnyFlow removes the need for the additional Causal ODE and Causal Consistency Distillation (CD) initialization stages used by Causal Forcing \citep{zhu2026causal} and Causal Forcing++ \citep{zhao2026causalforcingscalablefewstep}, respectively. The final DMD \citep{yin2024one} stage then trains on model-generated trajectories to align the causal generator with its inference-time distribution.

\subsection{Bidirectional Adaptation}
\label{sec:bidirectional_training}

Let $\mathbf{z}_0$ denote a clean video latent and let
$\mathbf{c}$ collect the text, image, and camera conditions.  Given a noisy
latent $\mathbf{z}_t$, the model predicts the native flow or velocity target
$\mathbf{u}_t$ of its backbone.  The first stage minimizes
\begin{equation}
  \mathcal{L}_{\mathrm{bid}}
  = \mathbb{E}_{\mathbf{z}_0,t,\boldsymbol{\epsilon}}
    \left[
      \left\|
        f_{\theta}(\mathbf{z}_t,t,\mathbf{c})-\mathbf{u}_t
      \right\|_2^2
    \right],
  \label{eq:bidirectional_objective}
\end{equation}
using bidirectional attention over the complete training window. This stage adapts the pretrained visual prior to camera-conditioned world data while retaining unrestricted temporal context. Camera control is injected through fused-PRoPE (Section~\ref{sec:fused_prope}): projective rotations act directly on the query, key, and value tensors in the existing self-attention path, requiring neither a separate control branch nor an additional attention pass. The resulting checkpoint provides both the initialization for causal training and the fixed bidirectional reference for DMD.

\subsection{Teacher-Forced AnyFlow Initialization}
\label{sec:anyflow_ar_initialization}

We next introduce causality while keeping the optimization target fully
supervised.  The latent sequence is divided into ordered blocks.  When
predicting the current block, the model can attend only to the current noisy
state and clean ground-truth history; future blocks are hidden.  For a sampled
pair of noise levels $(t,r)$, the training objective is
\begin{equation}
  \mathcal{L}_{\mathrm{TF\text{-}AF}}
  = \mathbb{E}
    \left[
      \sum_k
      \ell_{\mathrm{AF}}
      \left(
        f_{\phi};
        \mathbf{z}^{k}_{t},t,r,
        \mathbf{z}^{<k}_{0},\mathbf{c}
      \right)
    \right],
  \label{eq:teacher_forced_anyflow}
\end{equation}
where $\ell_{\mathrm{AF}}$ supervises the flow map between arbitrary noise
levels, directly enabling few-step autoregressive generation.

Initialized from the bidirectional model, this stage is a short causal
adaptation rather than a new round of representation learning.  The AnyFlow
objective exposes the model directly to the finite noise transitions used by
few-step sampling, while teacher forcing supplies a stable clean history.
Their combination yields a few-step autoregressive checkpoint that can be
passed directly to DMD, without a separate Causal ODE or Causal CD stage.

\subsection{DMD-based Causal Training}
\label{sec:dmd_causal_training}

Teacher-forced training still observes clean history, while inference
conditions on the model's own predictions.  We address this exposure gap with
DMD-based causal training.  A causal student generates trajectories under the
same temporal rule used at inference with a detached rollout-and-replay procedure \citep{zhuang2026selfgradientforcing} to allow KV Cache gradients. A frozen bidirectional teacher estimates
the target distribution, while a trainable fake-distribution model follows the
student's evolving rollout distribution.  Their difference provides the
distribution-matching direction for the student.

Writing $G_{\psi}$ for the causal generator, $s_{\mathrm{real}}$ for the
teacher score, and $s_{\mathrm{fake}}$ for the score of the generated
distribution, the DMD update takes the general form
\begin{equation}
  \nabla_{\psi}\mathcal{L}_{\mathrm{DMD}}
  = \mathbb{E}
    \left[
      J_{G_{\psi}}^{\top}
      w(t)
      \left(
        s_{\mathrm{fake}}(\mathbf{z}_t,t,\mathbf{c})
        -
        s_{\mathrm{real}}(\mathbf{z}_t,t,\mathbf{c})
      \right)
    \right],
  \label{eq:dmd_causal_objective}
\end{equation}
where $J_{G_{\psi}}$ maps the distribution-matching signal back to the
generator parameters and $w(t)$ controls its scale across noise levels.  The
fake-distribution model is optimized on generated samples, whereas the
bidirectional teacher remains fixed.  This separation preserves a stable
quality reference while allowing the training signal to track the changing
student distribution.
\section{Open Data Engine}
\label{sec:data}

The data component of \project is designed as an \emph{engine} rather than
as a fixed list of training clips.  Its central principle is to process and
annotate every canonical clip from every source before applying any
training-time selection criteria.  A clip that does not satisfy our released
default recipe is therefore retained in a separate rejected partition,
together with its annotations, metric values, and machine-readable rejection
reasons.  The released \texttt{high} and \texttt{xhigh} partitions are two
convenient quality tiers, not an irreversible definition of valid data.
Users can change thresholds, disable a filtering criterion, rebalance sources,
or construct a backbone-specific temporal view without rerunning the expensive
camera-estimation, captioning, and quality-assessment pipelines.

\begin{figure}[t]
  \centering
  \includegraphics[width=\textwidth]{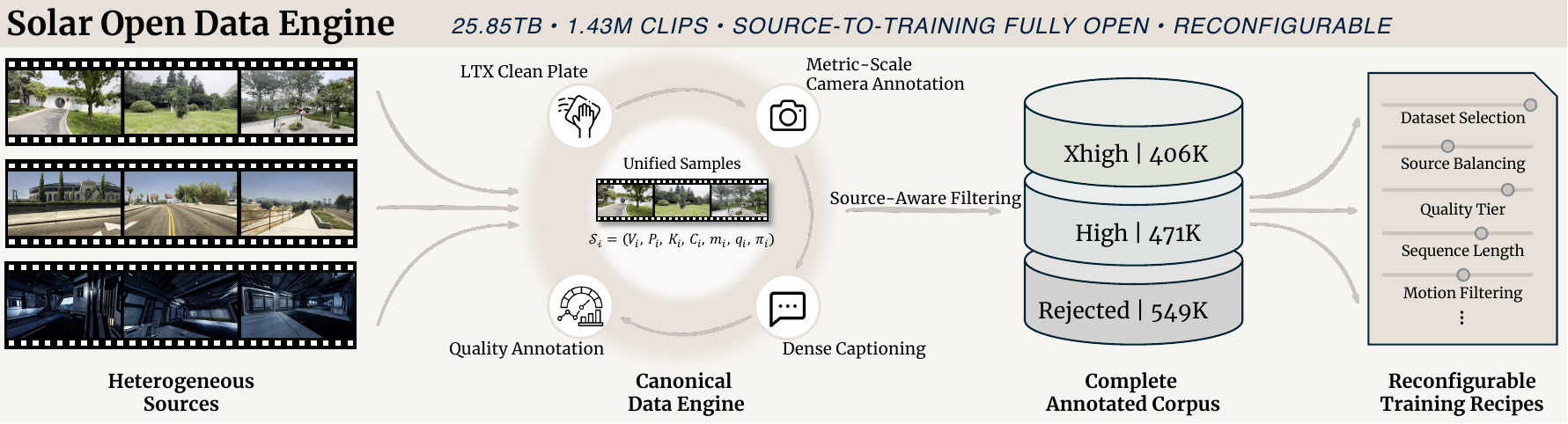}
  \caption{\textbf{Overview of the Solar open data engine.}}
  \label{fig:data-engine-overview}
\end{figure}

Our current corpus contains 1.43M canonical clips derived from 10 source datasets and organized into 14 independently addressable processing partitions, which we refer to as \emph{dataset owners}. The 10 source datasets are ABOT-World \citep{jiang2026abotworld}, DL3DV \citep{ling2023dl3dv}, MiraData \citep{ju2024miradata}, RealCam \citep{zheng2025realcam}, SpatialVID \citep{wang2025spatialvid}, Sekai-Game and Sekai-Walking \citep{li2025sekai}, MIND \citep{ye2026mind}, MultiCamVideo \citep{multicamvideo2026dataset}, and OmniWorld \citep{zhou2025omniworld}. At the owner level, DL3DV is divided into two temporal views, DL3DV-10s and DL3DV-60s, while three additional Clean Plate owners are derived from MiraData, Sekai-Walking, and SpatialVID. This organization yields 14 dataset owners in total, each of which can be filtered, weighted, and selected independently during training-recipe construction. As illustrated in \cref{fig:data-engine-overview}, the Solar Open Data Engine exposes the complete path from source ingestion to training-recipe construction. The remainder of this section describes the unified data contract, processing stages, selection policies, and resulting corpus.

\subsection{Unified Data Representation}
\label{sec:data-schema}

We normalize every source into a unified, frame-aligned sample representation.
Each sample is assigned a stable identifier, and all temporal modalities,
including video, camera poses, intrinsics, captions, and audio when available,
are aligned to the same, deterministically selected frame interval.

\paragraph{Unified sample schema.}
Every sample follows the same logical contract:
\begin{equation}
  \mathcal{S}_i =
  \bigl(V_i,\, P_i,\, K_i,\, C_i,\, m_i,\, q_i,\, \pi_i\bigr),
\end{equation}
where $V_i$ is the video, $P_i\in\mathbb{R}^{N\times4\times4}$ contains
metric camera-to-world transforms, $K_i\in\mathbb{R}^{N\times4}$ stores
per-frame $(f_x,f_y,c_x,c_y)$, $C_i$ is the dense caption, $m_i$ contains
source and media metadata, $q_i$ is the complete metric record, and $\pi_i$
stores provenance and lineage.  The schema records the source owner, original
identifier, temporal interval, processing versions, and the outcome of every
stage.  This makes samples from different sources interchangeable to the
training reader while keeping each derived artifact traceable to its input.

We separate three namespaces that are often conflated in dataset releases.
The \emph{physical corpus} contains the canonical samples and annotations; a
\emph{logical recipe} contains split membership, tier policy, source weights,
and repeat factors; and a \emph{model view} contains backbone-specific windows
or precomputed latents.  Changing a recipe therefore does not duplicate the
underlying videos, and changing a VAE does not alter the underlying data
selection.

\subsection{Metric-Scale Camera Annotation}
\label{sec:data-camera}

We follow the robust camera-annotation design of SANA-WM
\citep{zhu2026sanawm}, selecting the annotation path according to the geometry
available from each source.  For video-only sources, Pi3X estimates a
temporally consistent but scale-ambiguous scene structure, while MoGe-2
provides per-frame metric-depth anchors.  Their depths are aligned and fused
before being passed to a modified VIPE SLAM backend.  VIPE then estimates the
6-DoF camera trajectory and performs bundle adjustment with independently
optimizable per-frame $(f_x,f_y,c_x,c_y)$, initialized by GeoCalib.  This
per-frame formulation is important for clips containing zoom or focal-length
drift, for which a single intrinsic matrix for the entire video is
insufficient.

When a source provides ground-truth or COLMAP camera poses, we preserve the
supplied trajectory rather than replacing it with a SLAM estimate.  Pi3X is
used only to connect the predicted scene structure to the trajectory's metric
gauge: predicted and reference camera centers are aligned with a robust
Umeyama Sim(3) fit, which is re-estimated from the lowest-residual 80\% of
frames.  The pipeline also supports sources with metric depth but no camera
trajectory, using the depth directly as the scale anchor for VIPE.

All modes emit a frame-aligned metric camera-to-world (c2w) trajectory
$P_i\in\mathbb{R}^{N\times4\times4}$ and per-frame pixel-space intrinsics
$K_i\in\mathbb{R}^{N\times4}$.  Translations remain in metric units; no
corpus-wide translation normalization is applied.  Source-specific coordinate
conversion is performed once at ingestion and recorded in the sample metadata.
We require video frames, poses, and intrinsics to share exactly the same frame
indices.  Each sample is further checked for finite geometry, positive focal
lengths, field of view in a plausible range, focal consistency
$|f_x-f_y|/(\frac{1}{2}(f_x+f_y)+\epsilon)$, and temporal scale stability
$\operatorname{std}(s)/\operatorname{mean}(s)$.  Missing or non-finite camera
evidence fails closed whenever the corresponding recipe uses a camera gate.

\subsection{LTX Clean Plate Processing}
\label{sec:data-clean-plate}

\begin{figure}[t]
    \centering
    \includegraphics[width=\linewidth]{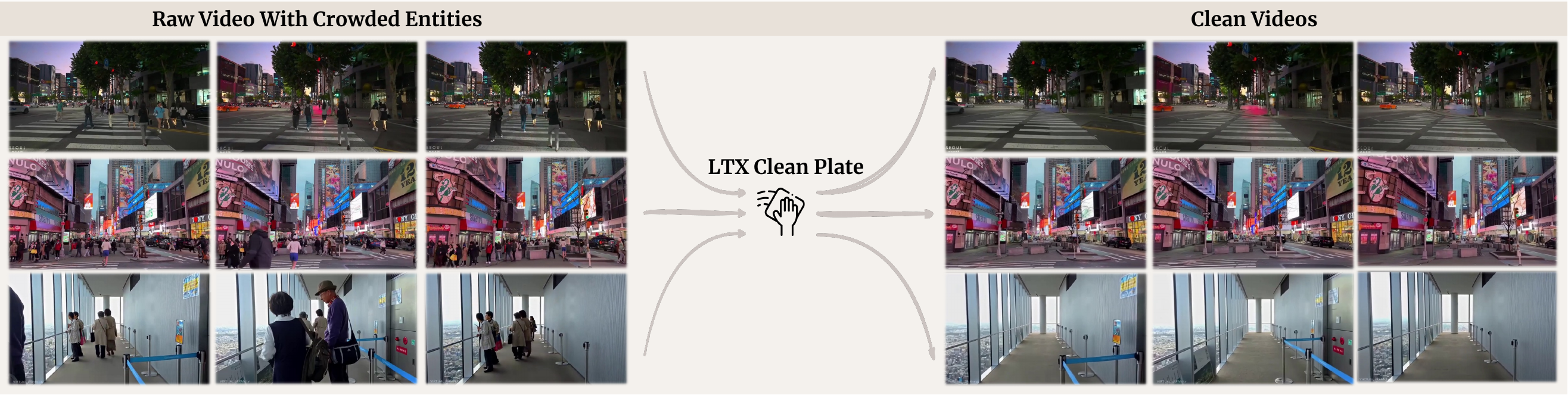}
    \caption{\textbf{LTX Clean Plate processing.} Representative frames before
    (left) and after (right) applying the LTX-2.3 Clean Plate pipeline.  The
    transformation removes dynamic people and vehicles while preserving the
    static scene layout and the source camera trajectory, yielding cleaner
    supervision for camera-controlled world-model training.}
    \label{fig:ltx-clean-plate}
\end{figure}

Dynamic people and vehicles are a frequent source of ambiguity for
camera-controlled world models: they introduce motion that is neither
explained by the camera nor consistently controllable.  We therefore produce
three additional Clean Plate derivative owners with the LTX-2.3 Clean Plate
IC-LoRA pipeline.  \Cref{fig:ltx-clean-plate} provides a visual comparison of
the source clips and their cleaned counterparts.
For each accepted source interval, the transformation removes people and
vehicles while retaining scene layout and camera motion.  We use eight denoising steps with strength $1.0$ and the Clean Plate prompt. Inputs are processed at $1248\!\times\!704$, and outputs are saved at $1280\!\times\!720$ and 16 fps.

Clean outputs are not assumed to inherit the quality of their source clips.
Removal can introduce texture artifacts, temporal discontinuities, or altered
motion statistics, so captions, visual metrics, semantic metrics, and camera
diagnostics are recomputed for every output window.  Metric scale is restored
from deterministic source-frame correspondences; invalid or ambiguous scale
reconstruction fails closed.  The current release contains 543k clean
clips: 298k SpatialVID-Clean clips (73k at 81 frames and 224k at
160 frames), 135k MiraData-Clean clips, and 109k
Sekai-Walking-Clean clips.  These are independent recipe owners, rather than
silent replacements for the original datasets.

\subsection{Dense Captioning and Multi-Axis Annotation}
\label{sec:data-annotation}

\paragraph{Kimi-K2.6 captions.}
All 1.43M canonical clips are captioned by the same Kimi-K2.6 production
pipeline.  Its prompt asks the model to inspect the entire clip and describe
only the persistent environment and stable scene content.  Dynamic entities,
actions, camera motion, shot terminology, and speculative or media-related
wording are excluded from the training caption, thereby reducing the risk that
the text condition leaks camera-control information.  The prompt and response
contract is summarized in \cref{tab:kimi-contract}.

\begin{table}[t]
\centering
\caption{\textbf{Kimi-K2.6 prompt and structured-output contract.}  The response
is a strict JSON object with exactly six top-level fields.}
\label{tab:kimi-contract}
\footnotesize
\renewcommand{\arraystretch}{1.2}
\begin{tabular}{P{0.20\textwidth} P{0.25\textwidth} P{0.49\textwidth}}
\hlineB{2.5}
\rowcolor{origintablehead}
\textbf{Item} & \textbf{Type or allowed values} &
\textbf{Prompt requirement and released representation} \\
\hlineB{1.5}
\rowcolor{origintablerow}
Visual input & 1 fps; at most 64 frames & Sample the entire clip with ffmpeg
round-up; maximum image edge 768 pixels, JPEG qscale 3, and no audio. \\
Caption target & One English paragraph; 60--150 words & Describe persistent
architecture, terrain, vegetation, water, furniture, materials, lighting,
weather, atmosphere, and stable spatial relations. \\
\rowcolor{origintablerow}
Caption exclusions & Hard negative instruction & Exclude people and animals,
dynamic or uncertain vehicles, actions, camera motion, shot/viewpoint terms,
media terminology, speculation, and invented details. \\
Response protocol & Strict JSON; no extra keys & Temperature 0, seed 0,
\texttt{thinking=false}; return only the JSON object, without Markdown or
explanatory text. \\
\hline
\rowcolor{origintablerow}
\texttt{dense\_caption} & String satisfying the caption contract & Becomes
\texttt{meta.caption} and \texttt{prompt.txt} (caption followed by one newline). \\
\texttt{vlm\_entity\_density} & Integer in $\{1,2,3\}$ & Peak combined density
of visible people, vehicles, and animals: none, sparse, or dense;
stored as \texttt{metrics.vlm\_entity\_density}. \\
\rowcolor{origintablerow}
\texttt{vlm\_quality} & Number in $\{1.0,2.0,3.0,4.0,5.0\}$ & Overall visual
usability from unusable to clean and richly describable; stored as
\texttt{metrics.vlm\_quality}. \\
\texttt{reject\_flags} & Deduplicated subset of eight flags & Allowed values are
\texttt{text\_heavy}, \texttt{watermark}, \texttt{ui\_overlay},
\texttt{blurry}, \texttt{near\_static}, \texttt{low\_light}, \texttt{nsfw},
and \texttt{single\_color}; stored as \texttt{metrics.vlm\_reject\_flags}. \\
\rowcolor{origintablerow}
\texttt{scene\_type} & \texttt{real\_world}, \texttt{rendered}, \texttt{game},
\texttt{animation}, or \texttt{mixed} & Classifies the visible world and is
stored as \texttt{metrics.vlm\_scene\_type}. \\
\texttt{scene\_transition} & Object with exactly \texttt{label},
\texttt{count}, \texttt{timestamps\_sec}, and \texttt{evidence} & The label is
\texttt{none}, \texttt{possible}, or \texttt{definite}; continuous fast motion
is not a transition.  Stored as \texttt{metrics.vlm\_scene\_transition}. \\
\hlineB{2.5}
\end{tabular}
\end{table}

Only \texttt{dense\_caption} becomes a text condition; the remaining fields are
stored as structured annotations and can be used by source-specific recipe
policies.  Transition predictions remain marked \texttt{unverified}, and
the presence of a caption alone does not affect tier assignment.

\paragraph{Quality and motion metrics.}
No single score captures the properties needed for interactive world-model
training.  We consequently retain a vector of complementary measurements:

\begin{itemize}
  \item \textbf{camera integrity:} finite pose/intrinsic tests, horizontal and
  vertical field of view, focal divergence, and scale coefficient of
  variation;
  \item \textbf{visual quality:} saturation statistics and the mean of DOVER's
  technical and aesthetic scores;
  \item \textbf{motion:} FFmpeg VMAF Motion and dense correspondence magnitude
  from UniMatch/GMFlow;
  \item \textbf{temporal consistency:} PySceneDetect cut count and the
  Kimi-K2.6 transition field; and
  \item \textbf{semantic suitability:} Kimi-K2.6 quality, entity density,
  scene type, and explicit flags such as \texttt{blurry},
  \texttt{ui\_overlay}, \texttt{watermark}, and \texttt{near\_static}.
\end{itemize}

All measurements, including ones unused by our default selection, remain in
the released record.  This is important because different source domains have
different natural score distributions: a universal motion or aesthetic
threshold would remove useful camera trajectories from one domain while
retaining artifacts in another.

\subsection{Source-Aware Filtering and Quality Tiers}
\label{sec:data-filtering}

We freeze a versioned policy per dataset owner.  The policy produces three
disjoint physical labels: \texttt{xhigh}, \texttt{high}, and
\texttt{rejected}.  A sample is \texttt{xhigh} if it satisfies both the kept
rule and the stricter promotion rule; otherwise, a sample satisfying the kept
rule is \texttt{high}.  All other samples are retained as \texttt{rejected}
with their failure reasons.  Every metric named by a rule must be present,
finite, and correctly typed.

For compactness, \cref{tab:data-policy} uses the following notation.  $N$ is
the frame count; $Q$ is Kimi-K2.6 quality; $R$ is its reject-flag set; $T$ is
the Kimi transition count; $E$ is entity density; $D$ is mean DOVER; $V$ is
VMAF Motion; $U$ is UniMatch motion; $S$ is mean saturation; and $J$ is the
PySceneDetect cut count.  $C_{120}$ requires both fields of view in
$[25,120]$ degrees, focal divergence $\leq0.20$, and scale coefficient of
variation $\leq2$; $C_{125}$ changes only the upper field-of-view bound to
125 degrees.  $G$ denotes the stricter post-Clean geometry and lineage check,
and $A$ denotes a non-empty caption.  All intervals are inclusive.  An
``xhigh extra'' is applied in addition to the kept rule.

\begin{table}[H]
\centering
\caption{\textbf{Frozen source-aware selection policies.}  The table is executable
policy documentation rather than a qualitative summary.}
\label{tab:data-policy}
\renewcommand{\arraystretch}{1.25}
\setlength{\tabcolsep}{4pt}
\resizebox{\textwidth}{!}{%
\begin{tabular}{P{3.9cm}|P{13.4cm}| P{4.4cm}}
\hlineB{2.5}
\rowcolor{origintablehead}
\textbf{Dataset owner} & \textbf{Kept rule} & \textbf{xhigh extra} \\
\hlineB{1.5}
\rowcolor{origintablerow}
DL3DV-10s & $C_{120}$; $S\!\in[0,180]$; $Q\!\geq4$; $D\!\geq.20$ & $Q=5$; $D\!\geq.30$ \\
DL3DV-60s & $C_{120}$; $S\!\in[0,180]$; $Q\!\geq4$; $D\!\geq.20$ & $Q=5$; $D\!\geq.30$ \\
\rowcolor{origintablerow}
MIND & $C_{125}$; $R=\varnothing$; $T=0$; $Q\!\geq4$ & $Q=5$ \\
ABOT & $C_{120}$; $S\!\in[0,180]$; $R=\varnothing$; $T=0$; $Q\!\geq4$ & $Q=5$ \\
\rowcolor{origintablerow}
OmniWorld & $C_{120}$; $S\!\in[0,180]$; $R=\varnothing$; $T=0$; $Q\!\geq4$; $D\!\geq.30$; $V\!\in[.5,100]$; $U\!\in[3,250]$ & $Q=5$; $U\!\leq200$ \\
Sekai-Game & $U\!\leq350$; $D\!\geq.20$; $Q\!\in[3,5]$; $\texttt{blurry}\notin R$ & $Q\!\in[4,5]$; $R=\varnothing$ \\
\rowcolor{origintablerow}
MiraData & $A$; $N\!\geq81$; $C_{120}$; $S\!\in[0,180]$; $Q\!\geq4$; $D\!\geq.40$; $\mathrm{resolution}=1280\!\times\!720$; $R=\varnothing$; $T=0$; $V\!\in[.5,50]$; $U\!\in[3,120]$ & $Q=5$; $U\!\leq100$ \\
RealCam & $N\!\geq81$; $C_{120}$; $S\!\in[0,180]$; $T=0$; $Q\!\geq4$; $V\!\in[.5,100]$; $D\!\geq.30$; $U\!\in[3,200]$; $\{\texttt{blurry},\texttt{ui\_overlay}\}\cap R=\varnothing$ & $Q=5$; $R\!\subseteq\!\{\texttt{near\_static}\}$ \\
\rowcolor{origintablerow}
MultiCamVideo & $C_{120}$; $S\!\in[0,180]$; $R=\varnothing$; $T=0$; $Q\!\geq4$; $D\!\geq.40$ & $Q=5$ \\
Sekai-Walking & $A$; $N\!\geq81$; $C_{120}$; $S\!\in[0,180]$; $T=0$; $Q\!\geq4$; $V\!\in[.5,50]$; $D\!\geq.35$; $U\!\in[3,120]$; $R\!\subseteq\!\{\texttt{watermark},\texttt{near\_static}\}$ & $Q=5$; $R=\varnothing$ \\
\rowcolor{origintablerow}
MiraData-Clean & exact kept-source lineage; $A$; $N\!\geq81$; $G$; $\mathrm{resolution}_{\mathrm{src}}=\mathrm{resolution}_{\mathrm{out}}=1280\!\times\!720$; $S\!\in[0,180]$; $Q\!\geq4$; $D\!\geq.40$; $V\!\in[.5,50]$; $U\!\in[3,120]$; $R=\varnothing$; $T=0$; $J\!\in[0,1]$ & $Q=5$; $D\!\geq.45$; $U\!\leq100$ \\
Sekai-Walking-Clean & $A$; $N\!\geq81$; $G$; $S\!\in[0,180]$; $T=0$; $Q\!\geq4$; $D\!\geq.35$; $U\!\in[3,120]$; $V\!\in[.5,50]$; $E\!\neq3$; $R\!\subseteq\!\{\texttt{watermark},\texttt{near\_static}\}$ & $Q=5$; $R=\varnothing$ \\
\rowcolor{origintablerow}
SpatialVID & $A$; $N\!\geq81$; $C_{120}$; $S\!\in[0,180]$; $T=0$; $Q\!\geq4$; $V\!\in[.5,50]$; $D\!\geq.35$; $U\!\in[3,120]$; $J=0$; $R\!\subseteq\!\{\texttt{watermark},\texttt{near\_static}\}$ & $Q=5$; $R=\varnothing$ \\
SpatialVID-Clean & $A$; $N\!\geq81$; $G$; $S\!\in[0,180]$; $T=0$; $Q\!\geq4$; $V\!\in[.5,50]$; $D\!\geq.35$; $U\!\in[3,120]$; $J=0$; $E\!\neq3$; $R\!\subseteq\!\{\texttt{watermark},\texttt{near\_static}\}$ & $Q=5$; $R=\varnothing$ \\
\hlineB{2.5}
\end{tabular}%
}
\end{table}

The table intentionally contains asymmetric policies.  For example,
Sekai-Game preserves high-motion game trajectories without imposing the
generic camera-geometry or saturation gates, whereas clean owners require
post-transformation geometry checks.  Likewise, some metrics are retained only
as annotations for a given source.  This avoids retroactively presenting every
computed metric as a selection gate.  It also distinguishes the current
Kimi-K2.6 tier policy from earlier internal recipes that used different
semantic-score scales or fewer visual gates.

\subsection{Open Recipes and Split Construction}
\label{sec:data-recipes}

The physical WebDataset release stores every canonical row under one of
\texttt{kept-high}, \texttt{kept-xhigh}, or \texttt{rejected}, while recipe
indices provide logical training and evaluation membership.  Splits are
constructed using stable sample identifiers and are verified to contain no
overlapping identifiers.  We also publish independent 100-sample test views
for each owner (1,400 rows in total) for data inspection and reader validation.

A recipe may reference a row repeatedly without copying it.  For example,
our 81f short recipe spans ten owners and contains 600,320 physical training
rows, but source balancing applies a repeat factor of six to ABOT, MiraData,
and Sekai-Game, producing 870,210 virtual occurrences per epoch; its held-out
test view contains 1,000 rows.  Because selection, balancing, and
temporalization are expressed in indices, users can audit the released recipe
or create alternatives such as a metric-camera-only subset, less aggressive
motion filtering, a different source mixture, or a longer temporal curriculum.
Rejected rows remain useful
for such experiments because they carry the same annotations and explicit
rejection provenance as kept rows.

\subsection{Final Corpus and Temporal Views}
\label{sec:data-statistics}

\begin{figure}[t]
\begin{minipage}[t]{0.53\textwidth}
\centering
\captionsetup{type=table}
\caption{\textbf{Canonical clip inventory by dataset owner.}  High and xhigh
are disjoint tiers; kept is their sum.}
\label{tab:data-inventory}
\renewcommand{\arraystretch}{1.18}
\scriptsize
\setlength{\tabcolsep}{2.5pt}
\resizebox{\linewidth}{!}{%
\begin{tabular}{P{2.6cm}|N{1.6cm}N{1.6cm}N{1.6cm}N{1.6cm}}
\hlineB{2.5}
\rowcolor{origintablehead}
\textbf{Dataset owner} & \textbf{All} & \textbf{High} & \textbf{xhigh} &
\textbf{Rejected} \\
\hlineB{1.5}
\rowcolor{origintablerow}
ABOT & 30,966 & 127 & 30,715 & 124 \\
DL3DV-10s & 120,924 & 54,528 & 60,396 & 6,000 \\
\rowcolor{origintablerow}
DL3DV-60s & 10,077 & 3,578 & 6,065 & 434 \\
MIND & 533 & 117 & 402 & 14 \\
\rowcolor{origintablerow}
MiraData & 140,877 & 3,683 & 17,806 & 119,388 \\
MiraData-Clean & 135,224 & 12,865 & 5,740 & 116,619 \\
\rowcolor{origintablerow}
MultiCamVideo & 123,117 & 89,587 & 5,369 & 28,161 \\
OmniWorld & 19,632 & 3,773 & 13,552 & 2,307 \\
\rowcolor{origintablerow}
RealCam & 45,697 & 16,154 & 16,074 & 13,469 \\
Sekai-Game & 2,550 & 537 & 1,410 & 603 \\
\rowcolor{origintablerow}
Sekai-Walking & 22,990 & 6,054 & 12,976 & 3,960 \\
Sekai-Walking-Clean & 109,248 & 46,831 & 33,660 & 28,757 \\
\rowcolor{origintablerow}
SpatialVID & 365,345 & 100,815 & 127,180 & 137,350 \\
SpatialVID-Clean & 298,514 & 133,149 & 73,450 & 91,915 \\
\hline
\rowcolor{originaccent!12!white}
\textbf{Total} & \textbf{1,425,694} & \textbf{471,798} & \textbf{404,795} &
\textbf{549,101} \\
\hlineB{2.5}
\end{tabular}
}
\end{minipage}\hfill
\begin{minipage}[t]{0.44\textwidth}
\centering
\captionsetup{type=figure}
\caption{\textbf{Distribution of canonical clips by clip length and quality
tier.}}
\label{fig:data-frame-distribution}
\vspace{-1.3mm}
\includegraphics[width=\linewidth]{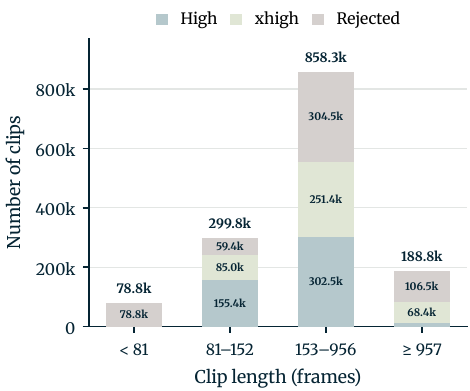}
\end{minipage}
\end{figure}

\Cref{tab:data-inventory,fig:data-frame-distribution} summarize the released
corpus from complementary source and temporal perspectives.  The owner-level
accounting in \cref{tab:data-inventory} shows that 876k of the 1.4M canonical clips are retained: 471k in the \texttt{high} tier and 404k in
\texttt{xhigh}.  The remaining 549k fully processed clips are published in
the rejected partition.  In physical storage, the complete corpus comprises
29k shards and approximately 25.85 TB.

\Cref{fig:data-frame-distribution} instead aggregates the same corpus by raw
clip length, making the tier composition of each temporal range directly
visible.  The majority of clips fall in the 153--956-frame interval, which
contains 858k clips overall and 553k kept clips.  All 78k clips shorter
than 81 frames belong to the rejected partition, whereas the long tail of clips
with at least 957 frames contains 82k kept clips.

From these physical clips, we construct model-ready temporal views for short- and long-horizon generation. The release separates the annotated source corpus from temporal windowing, data-mixture policies, and backbone-specific latent representations, with each layer indexed reproducibly. This design allows researchers to adapt temporal coverage, filtering criteria, source mixtures, and backbone-specific requirements without rebuilding the underlying corpus.

\section{\project Model Family}
\label{sec:model_family}

\project is organized as one interactive world-model framework rather
than as four isolated model releases.  The framework fixes a common data
recipe, a shared processing and annotation contract, a camera-geometry
interface, and a consistent training and inference workflow.  Different
video-generation backbones are then instantiated under this interface so
that their behavior can be studied under comparable data and system
conditions.  The shared interface does not erase backbone-specific
semantics: each route retains the temporal representation, attention
layout, conditioning details, and native optimization objective required
by its base model.

The resulting family contains four interactive video world models:
\project-wan2.2-5B, \project-wan2.2-14B, \project-ltx-2.5-22B, and
\project-minimax-h3-33B.  They cover four independent video-generation
routes---Wan 2.2-5B, Wan 2.2-14B, LTX-2.5, and MiniMax-H3---under one \project contract.
This organization makes two comparisons possible at the same time.  The
common data and camera interface isolates differences between backbones,
while the route-specific adapters preserve the mechanisms that make each
backbone usable for interactive generation.

\subsection{Unified Camera Conditioning with Fused-PRoPE}
\label{sec:fused_prope}

Camera control is a shared requirement of the \project family.  Each
camera trajectory is converted into a geometric condition aligned with the
video frames and injected into the video attention computation.  This
condition contains the camera poses, intrinsics, frame alignment, and the
associated normalization transforms.  A route adapter maps this common
condition into the representation expected by each backbone while
preserving the same underlying camera motion.

\project uses fused projective rotary positional embeddings (fused-PRoPE),
as in MosaicMem \citep{yu2026mosaicmem}.  At a high level, the procedure is the same for all
four models.  The backbone first applies its native video RoPE to represent
the temporal and spatial positions of video tokens.  Camera poses and
intrinsics then determine projective rotations that are applied to the
query, key, and value tensors.  A single self-attention operation follows,
after which the matching output transform is applied before the native
output projection.  Camera motion is therefore introduced through the
attention computation itself, rather than by appending an unrelated
condition token or a separate camera branch.
This procedure is shared across all four routes; backbone-specific details
are deferred to their respective paragraphs below.

\subsection{Comparison of the Four \project Models}
\label{sec:model_comparison}

The four model routes and their backbone-specific parameter, conditioning, and
latent-compression contracts are summarized in
\cref{tab:origin-model-family}.

\begin{table*}[t]
\centering
\caption{\textbf{Backbone-level differences within the \project model family.}
All routes share the same data, camera, and training/inference interfaces while
retaining backbone-specific parameter and conditioning contracts.}
\label{tab:origin-model-family}
\footnotesize
\renewcommand{\arraystretch}{1.25}
\setlength{\tabcolsep}{3pt}
\hyphenpenalty=10000
\exhyphenpenalty=10000
\resizebox{\textwidth}{!}{%
\begin{tabular}{M{2.65cm} M{3.35cm} M{4.20cm} M{4.1cm} N{3.20cm}}
\hlineB{2.5}
\rowcolor{origintablehead}
{\centering\bfseries Model\par} &
{\centering\bfseries Retained Components\par} &
{\centering\bfseries Image Conditioning\par} &
{\centering\bfseries Audio Handling\par} &
{\centering\bfseries VAE DS (T / S)\par} \\
\hlineB{1.5}
\rowcolor{origintablerow}
\makecell[l]{\mbox{\project-}\\\mbox{wan2.2-5B}} &
5B &
Native TI2V latent path; no separate $y$ &
None &
\makecell[c]{$4\times$ / $16\times$} \\
\makecell[l]{\mbox{\project-}\\\mbox{wan2.2-14B}} &
High-noise expert; full-timestep route &
Official $y$: 4-channel mask + 16-channel image latent &
None &
\makecell[c]{$4\times$ / $8\times$} \\
\rowcolor{origintablerow}
\makecell[l]{\mbox{\project-}\\\mbox{ltx-2.5-22B}} &
14.74B video path; 1.61B Gemma connector frozen &
Native first-frame latent &
3.69B audio stream and 2.58B AV cross-attention removed &
\makecell[c]{$8\times$ / $32\times$} \\
\makecell[l]{\mbox{\project-}\\\mbox{minimax-h3-33B}} &
33B Omni Transformer &
Qwen image/caption rows + VisualVAE first-image anchor &
Audio target rows retained; audio loss disabled &
\makecell[c]{$17n+5$ $\mapsto 5n+2$ \\ / $16\times$} \\
\hlineB{2.5}
\end{tabular}%
}
\end{table*}

The common \project interface is designed to preserve, rather than
replace, the capabilities of each video-generation base.  We therefore
retain each backbone's native temporal grid, text representation, attention
mask, training objective, and other settings wherever they are essential
to its behavior, while exposing them through the same data and camera
contracts.  The following paragraphs summarize these backbone-specific
choices before we introduce the shared training methodology.

\paragraph{\project-wan2.2-5B} is the TI2V member of the family.  The model
represents a clip with a 48-channel Wan2.2 VAE latent and does not take
a separate image condition tensor \(y\).  \project keeps this native interface:
the first latent is used as the clean image anchor and is excluded from the
video loss, while the remaining latents are denoised as video targets under
the shared camera condition.

\paragraph{\project-wan2.2-14B} starts from the Wan 2.2 I2V-A14B model, whose released
checkpoint contains separate high-noise and low-noise experts.  \project uses
the high-noise expert for initialization and trains it as one dense model
over all timesteps; the two-expert routing rule and its noise boundary are
not retained.  Its image interface is also distinct from the 5B route.  We
construct the official \(y\) by encoding a video whose first frame is the
input image and whose later frames are zero, then concatenate a four-channel
first-frame mask and the resulting 16-channel image latent to the noisy
16-channel target.  Thus \(y\) is an explicit input, and the first target
latent remains noisy and supervised.

\paragraph{\project-ltx-2.5-22B} is released as an audio-visual checkpoint, but \project uses it
as a video-only backbone.  The audited checkpoint assigns 3.7B parameters
to the audio stream and 2.6B to bidirectional audio--video cross-attention;
both components are removed.  \project retains the 13.1B video core and
loads the 1.6B Gemma video connector frozen, giving a 14.7B retained
video route driven by cached Gemma4 features.  The route keeps LTX's native
first-frame latent convention and rectified-flow objective, and inserts the
shared camera condition only in video self-attention.

\paragraph{\project-minimax-h3-33B} keeps MiniMax-H3's multimodal packing rather than
turning H3 into a video-only DiT.  Each packed sequence contains Qwen
caption/image conditions, a one-frame VisualVAE image anchor, audio rows,
and target-video rows.  The audio rows are retained because they are small;
they are filled with an official AudioVAE encoding of silence and assigned
zero audio loss, so only target-video rows are supervised.  The Qwen
conditioner, VisualVAE, and AudioVAE are frozen, while the H3 Omni
Transformer is trained with its native temporal, masking, and camera
semantics.  Its camera injection also preserves the pretrained head split:
dimensions $[0{:}96)$ retain the native content representation and MM-RoPE,
whereas only dimensions $[96{:}128)$ receive camera-relative-pose PRoPE on
the VisualVAE anchor and target-video rows.

\section{Experiments}
\label{sec:experiments}

We evaluate two stages of the \project training pipeline.  We
first examine the \emph{bidirectional pretrained models} across all four
backbone routes on 10-second OOD generations initialized from images
created with GPT Image 2 or Krea, testing how consistently the shared data and
camera contract transfers across heterogeneous video generators and unseen
visual domains.  We then examine \emph{distilled causal generation} under both
in-domain and OOD settings, spanning 10-second, minute-scale, and hour-scale
horizons.  Together, these settings expose model behavior across heterogeneous
backbones, input domains, training stages, and rollout horizons.

\subsection{Configurations}
\label{sec:configurations}

We use checkpoints produced by the training pipeline in
\cref{sec:training}.  All
reported \project causal results are generated at 16 fps with
four sampling steps and no attention sink.  Every sequence is initialized from
one image and conditioned on a text description and a frame-aligned camera
trajectory.  Bidirectional generation produces 10-second clips with
bidirectional temporal attention.  Causal generation instead conditions each new unit on model-generated history to reach the reported 10-second, minute-scale, and
hour-scale horizons.  The scene prompt remains fixed throughout each sequence,
while the prescribed camera trajectory provides the only time-varying external
control.  We use the native latent representation, conditioning interface, and
sampling objective of each backbone, as described in \cref{sec:model_family}.

\subsection{Bidirectional Pretrained Models}
\label{sec:bidirectional_results}

We begin with the model family after bidirectional pretraining and before
causal adaptation or distillation.  Each model receives an OOD first frame
created with GPT Image 2 or Krea, a fixed scene caption, and a frame-aligned
camera trajectory, and generates the remaining frames.
Evaluating this stage separately tests whether the unified data and
camera-conditioning contract generalizes beyond the training domains while
preserving the visual prior of each native backbone.

\begin{figure*}[t]
  \centering
  \includegraphics[width=1\linewidth]{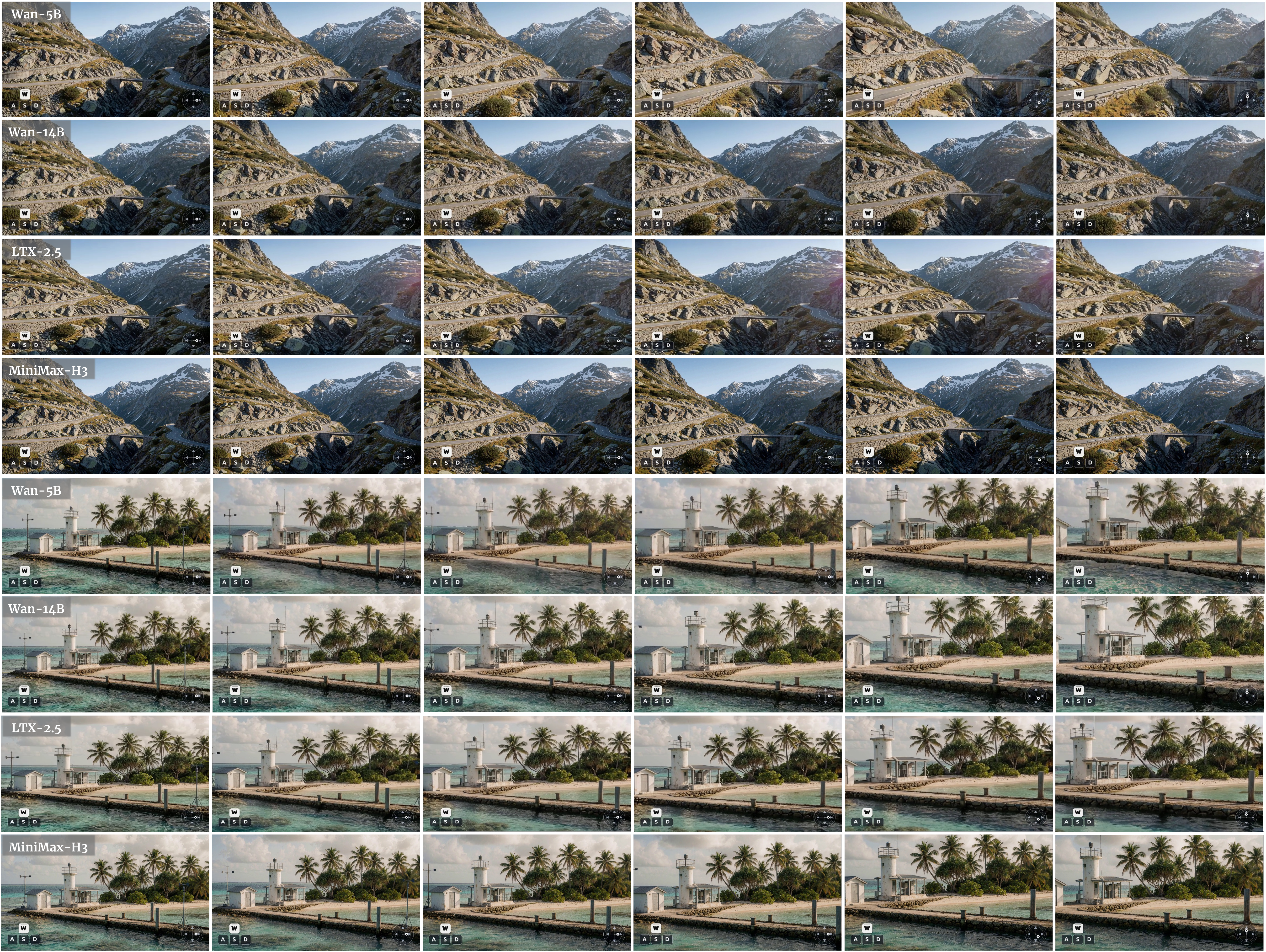}
  \caption{\textbf{OOD qualitative results of the bidirectional pretrained
  model family.}  Each backbone route generates a 10-second video from an OOD
  first frame and a prescribed camera trajectory.}
  \label{fig:bidirectional_qualitative}
\end{figure*}

As shown in \cref{fig:bidirectional_qualitative}, the four backbone routes are
evaluated on representative OOD inputs under translation, rotation, and mixed
6-DoF camera paths.

\subsection{Distilled Causal Generation}
\label{sec:causal_results}

We next examine distilled causal generation across familiar held-out inputs,
externally synthesized initial images, and increasingly long autoregressive
horizons.

\subsubsection{In-Domain Validation Results}
\label{sec:indomain_results}

We draw the in-domain evaluation examples from a held-out
validation pool. The pool is in-domain in the operational
sense that its samples are drawn from the same source families and processed by
the same canonical pipeline as the training mixture, while no validation
sample is used for optimization.

\begin{figure*}[t]
  \centering
  \includegraphics[width=1\linewidth]{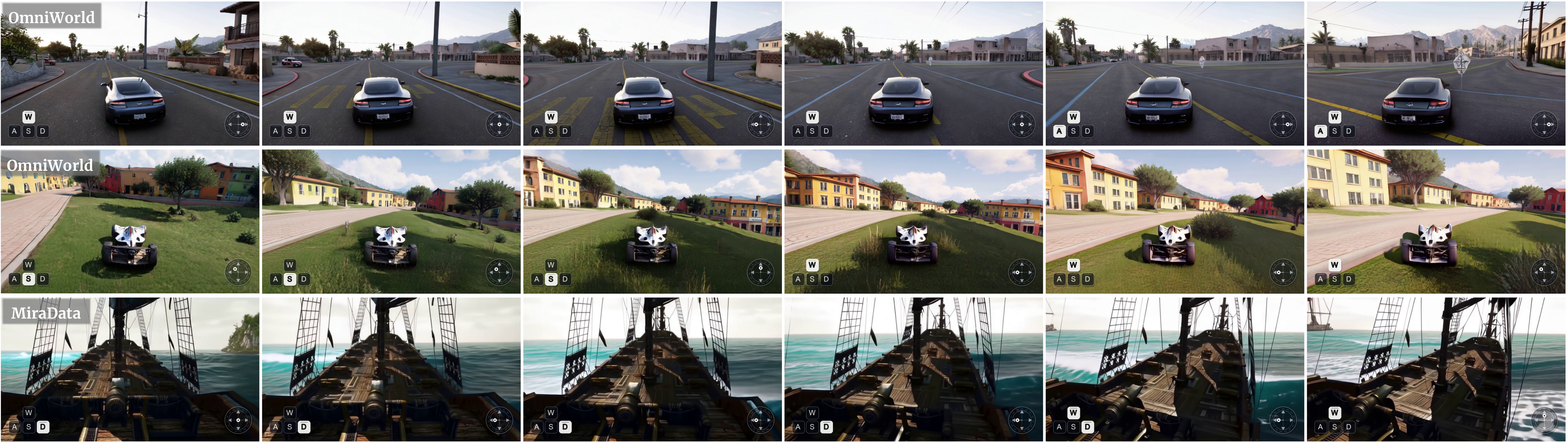}
  \caption{\textbf{In-domain third-person qualitative results
  generated by the \project-wan2.2-5B-fast causal student.}  Each sequence
  is initialized from a real first frame and follows a prescribed camera
  trajectory.  All subsequent frames are generated autoregressively, testing
  subject consistency and camera response from an external viewpoint.}
  \label{fig:indomain_qualitative_10s}
\end{figure*}

\begin{figure*}[t]
  \centering
  \includegraphics[width=1\linewidth]{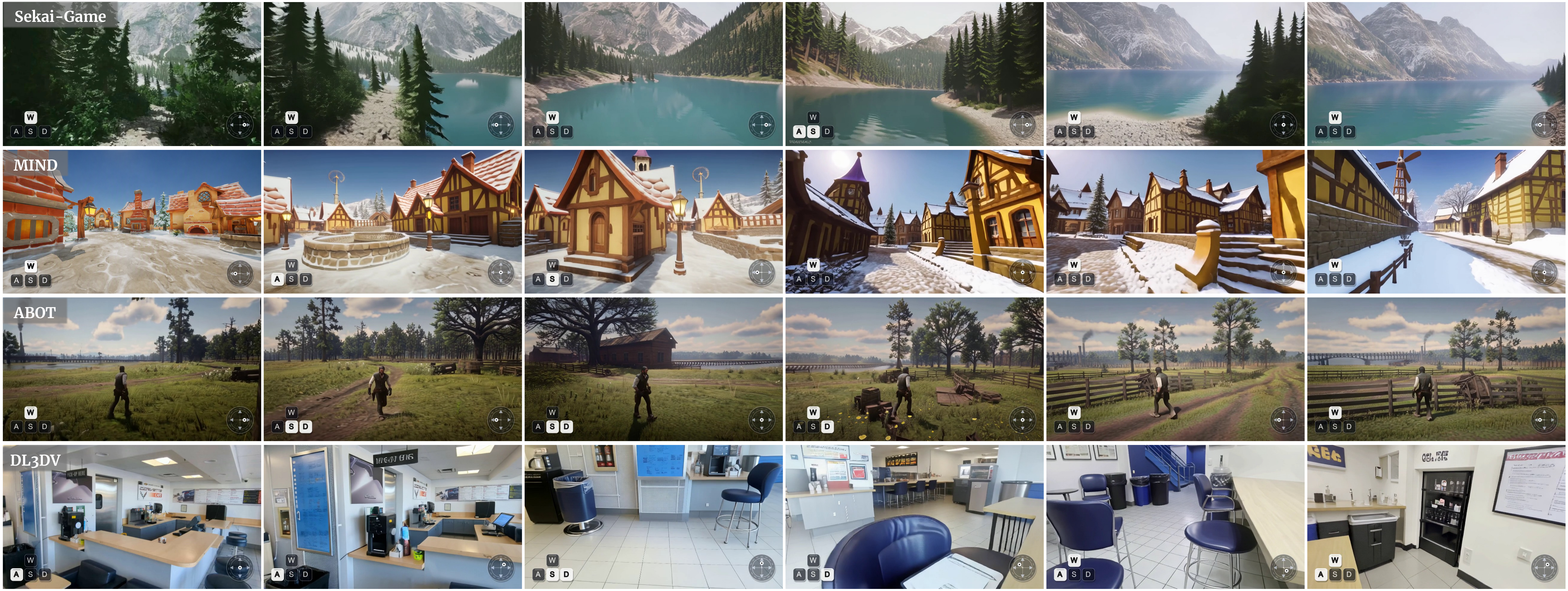}
  \caption{\textbf{In-domain minute-scale qualitative results generated by the
  \project-wan2.2-5B-fast causal student.}  Each uninterrupted sequence is
  initialized from its real first frame and follows a given camera trajectory;
  all subsequent frames are generated autoregressively.}
  \label{fig:indomain_qualitative}
\end{figure*}

For the selected examples, we report two complementary settings.  First,
\cref{fig:indomain_qualitative_10s} presents 10-second third-person rollouts in
which a visible subject is observed from an external viewpoint.  These examples
test whether the model can follow the prescribed camera trajectory while
maintaining the subject's appearance and its spatial relationship to the
surrounding scene.  Second, \cref{fig:indomain_qualitative} presents
uninterrupted minute-scale rollouts spanning representative real-world, synthetic,
and game-domain scenes under translation, rotation, and mixed 6-DoF camera
paths.

\subsubsection{Out-of-Domain Initialization}
\label{sec:ood_results}

A useful interactive world model should not require every explorable world to
originate from its training collection.  We therefore construct a separate
stress test in which each first frame is synthesized with GPT Image 2 or Krea.
The prompts deliberately cover visual conditions uncommon in the curated
corpus, such as stylized illustration, surreal geometry, unusual material
combinations, dramatic lighting, and imaginative indoor or outdoor
environments.  

For every initial image, we apply several camera trajectories,
including forward motion, lateral translation, turning, orbit-like motion, and
compositions of translation and rotation.  Using multiple trajectories from
the same image separates robustness to appearance shift from responsiveness to
the control signal.  We first report 10-second OOD rollouts in this subsection;
the minute-scale OOD stress test is presented in
\cref{sec:long_video_results}.

\begin{figure*}[t]
  \centering
  \includegraphics[width=\textwidth]{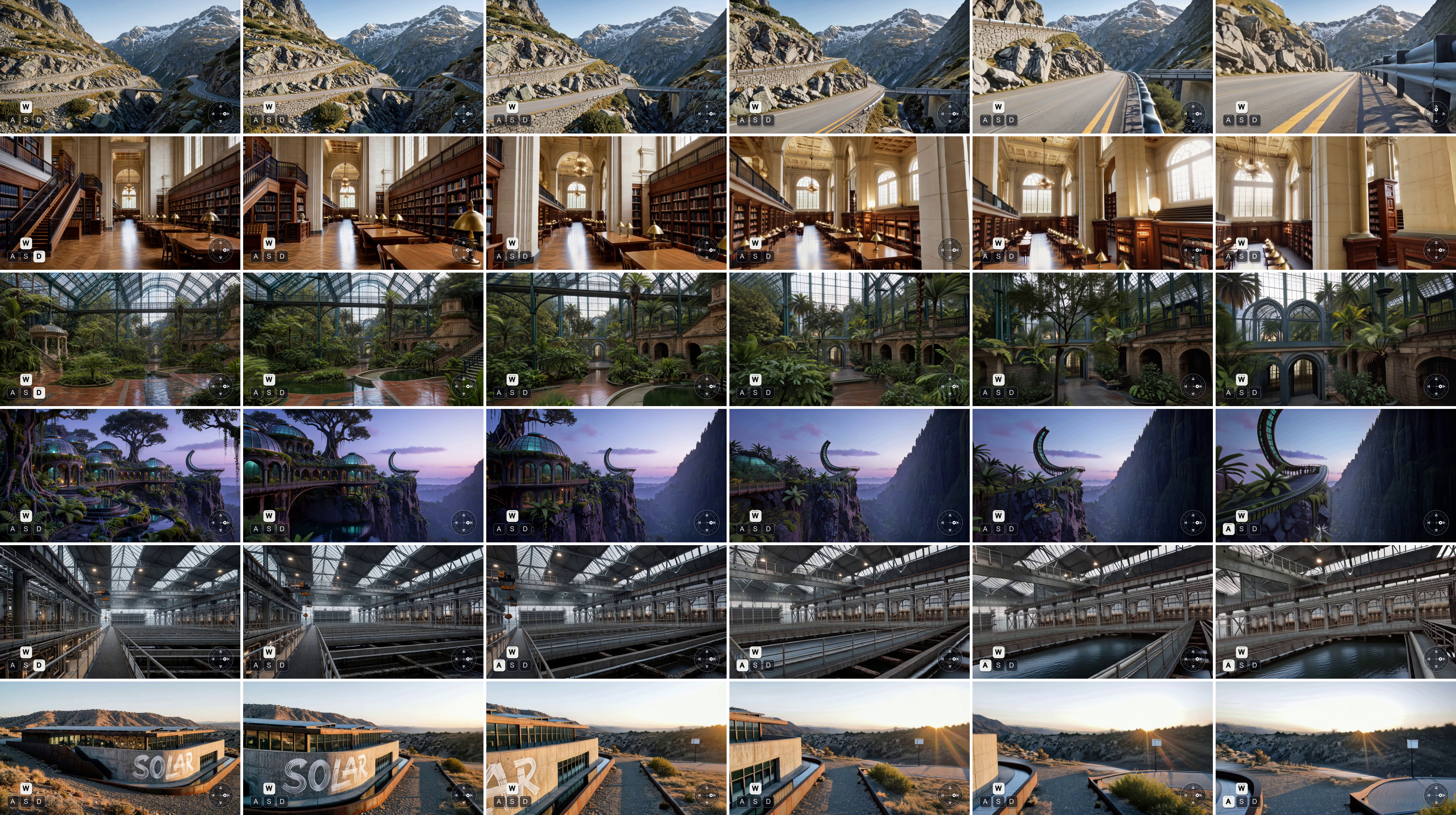}
  \caption{\textbf{OOD generalization from externally synthesized
  initial images, generated by the \project-wan2.2-5B-fast causal student.}
  Initial frames are paired with independently specified camera trajectories.
  Only the first image is externally provided; every subsequent frame in each
  10-second rollout is generated autoregressively.  The examples span photorealistic, stylized, and
  deliberately unusual scenes outside the released validation set.}
  \label{fig:ood_qualitative}
\end{figure*}

As illustrated in \cref{fig:ood_qualitative}, the model turns a broad range of
externally synthesized still images into navigable video while largely
retaining the initial style, palette, and dominant scene elements.  The
examples are particularly informative when camera motion reveals regions that
were not visible in the input: all newly exposed content must be inferred by
the world model, rather than copied from the image generator.  

\subsubsection{Long Video Generation}
\label{sec:long_video_results}

We evaluate continuous autoregressive generation at minute-scale and hour-scale
horizons.  Each long video starts from a single image, keeps the scene prompt
fixed, and follows a predetermined camera trajectory throughout the complete
run.  We do not restart from the input image, inject reference frames,
independently generate and splice short clips, or use an attention sink.

\paragraph{Minute-scale rollouts.}
We evaluate uninterrupted minute-scale generation from
synthesized first frames under camera trajectories that combine forward motion,
turning, lateral translation, and pauses.  These compound trajectories test
both sustained response to a control and transitions between successive
camera commands without resetting the model.

\begin{figure*}[t]
  \centering
  \includegraphics[width=\textwidth]{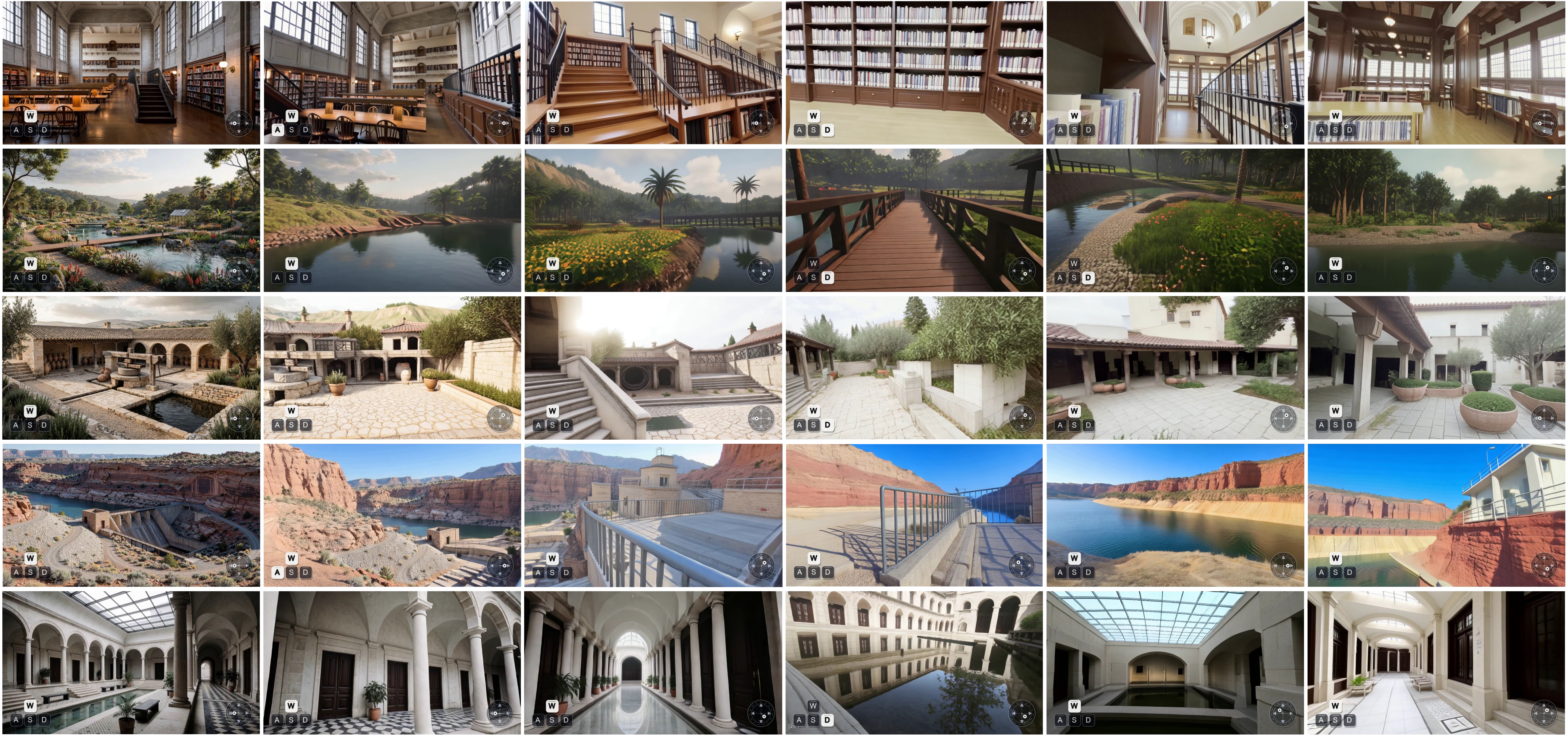}
  \caption{\textbf{Minute-scale OOD autoregressive rollouts generated by the
  \project-wan2.2-5B-fast causal student.}  Frames are sampled from uninterrupted
  generations initialized by a single OOD image.  The scene prompt remains
  fixed for the entire run.}
  \label{fig:minute_qualitative}
\end{figure*}

As shown in \cref{fig:minute_qualitative}, the model navigates five visually
distinct scenes while preserving their principal
layout and appearance.  Across the sampled timelines, newly revealed stairs,
rooms, vegetation, terraces, cliffs, and waterways remain compatible with the
initial observation, and the views continue to evolve with the prescribed
camera motion throughout the minute-scale rollout.  These results demonstrate
OOD appearance generalization and sustained camera control well beyond the
5-second sequences used during training.

\begin{figure*}[!t]
  \centering
  \includegraphics[width=\textwidth]{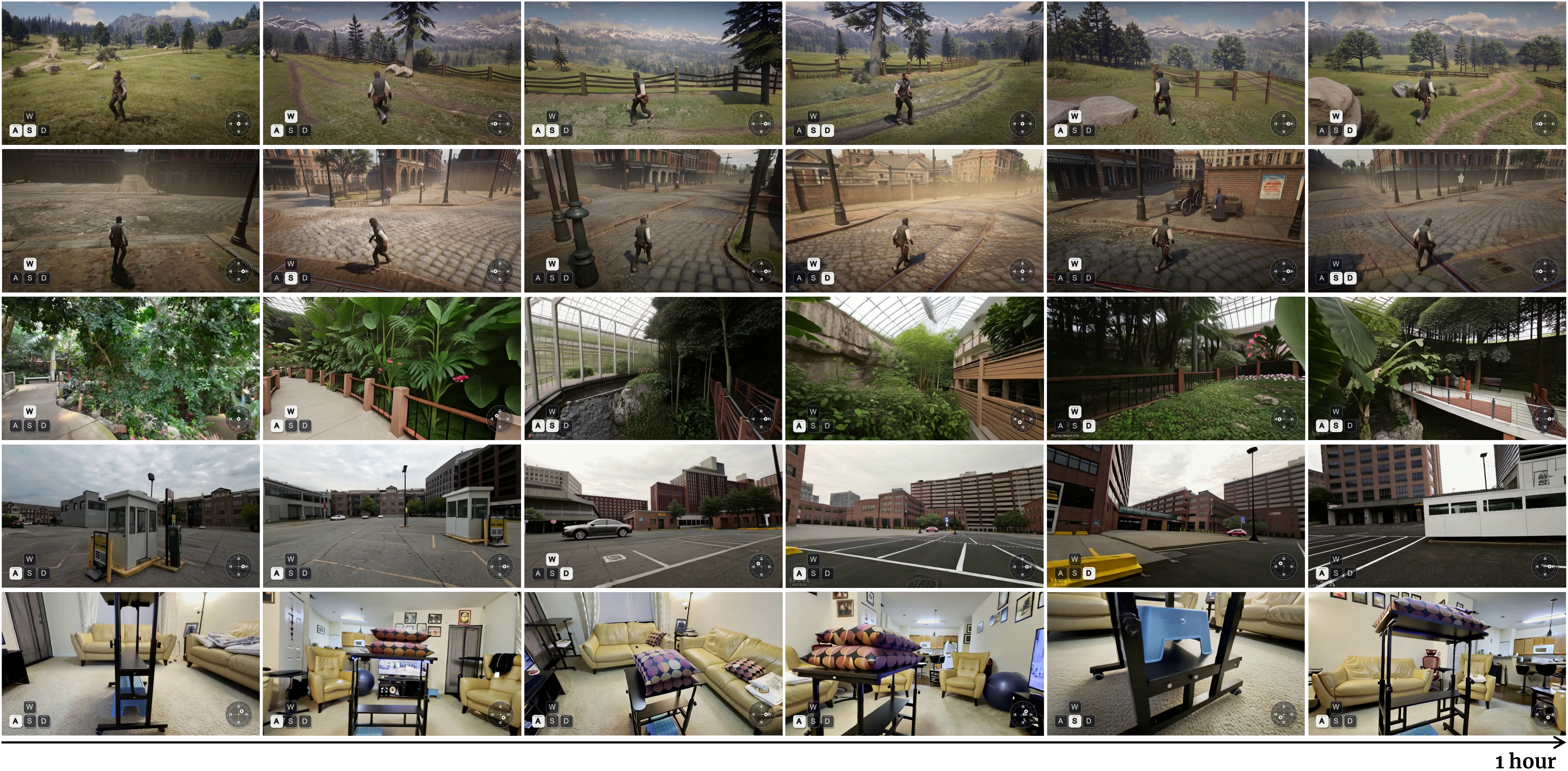}
  \caption{\textbf{Hour-scale world rollout generated by the
  \project-wan2.2-5B-fast causal student.}  Sparse frames are sampled throughout a single
uninterrupted autoregressive session initialized with a real first frame from
the held-out validation pool, spanning the initial observation to the 60-minute
endpoint. The scene prompt remains fixed and the model follows a
predetermined camera trajectory.}
  \label{fig:hour_qualitative}
\end{figure*}

\paragraph{Hour-scale rollouts.}
We further evaluate uninterrupted hour-scale
generation from real first frames selected from the held-out validation pool.
The model is trained only on 5-second sequences. During evaluation, it follows the
given camera trajectory. Consequently, all later frames are generated
autoregressively by repeatedly conditioning on model-generated history, making
the generation substantially longer than the training horizon.

As shown in \cref{fig:hour_qualitative}, the model sustains diverse scenes through uninterrupted hour-scale rollouts. Across the
sampled timeline, the generated views continue to respond to camera control
while retaining the principal scene layout and visual identity established by
the initial observation. The 60-minute endpoints remain recognizable and
visually coherent.

\section{Conclusion}
\label{sec:conclusion}

We presented \project, an open foundation for building interactive video world
models across heterogeneous data sources and video backbones.  \project
combines a reconfigurable multi-source data engine with backbone-native
adaptation and a unified three-stage training recipe, yielding four
camera-controllable models spanning 5B--33B parameters.  The resulting causal
models support real-time interaction and continuous hour-scale rollouts after
training on only 5-second sequences.  By releasing the data, processing
pipeline, training recipes, model weights, and training \& inference
framework, we hope to provide a reproducible and extensible basis for future world-model research.

\bibliography{main}

\end{document}